\documentclass[10pt,twocolumn,letterpaper]{article}
\usepackage[pagenumbers]{cvpr}

\usepackage{colortbl}
\definecolor{MaDeSRow}{HTML}{E4EDF8}

\definecolor{MaDeSGain}{HTML}{AEC9EA}
\newcommand{\MaDeSMaxGain}{22.3}
\newcommand{\gaincell}[2]{%
  \expanded{\noexpand\cellcolor{MaDeSGain!\fpeval{100*max(0,min(1,(#2)/\MaDeSMaxGain))}!white}}%
  #1 \gain{#2}%
}

\definecolor{cvprblue}{rgb}{0.21,0.49,0.74}
\usepackage[pagebackref,breaklinks,colorlinks,allcolors=cvprblue]{hyperref}
\usepackage{bm,multirow,algorithm,algorithmic}
\newcommand{\gain}[1]{\scriptsize(#1)}
\hypersetup{
  pdftitle={Two-Axis Event Dataset Distillation with Maturity-Decoupled Supervision},
  pdfauthor={Mingzhuo Li, Guang Li, Linfeng Ye, Jiafeng Mao, Takahiro Ogawa, Konstantinos N. Plataniotis, Miki Haseyama}
}
\title{Two-Axis Event Dataset Distillation with Maturity-Decoupled Supervision}
\author{
  Mingzhuo Li$^{1}$\quad Guang Li$^{1}$\thanks{Correspondence to Guang Li
  \textless{}guang@lmd.ist.hokudai.ac.jp\textgreater{}.}\quad Linfeng Ye$^{2}$\quad Jiafeng Mao$^{3}$\\
  Takahiro Ogawa$^{1}$\quad Konstantinos N. Plataniotis$^{2}$\quad Miki Haseyama$^{1}$\\[4pt]
  {\normalsize $^{1}$Hokkaido University\quad $^{2}$University of Toronto}\quad {\normalsize $^{3}$University of Tokyo}
}
\date{}
\begin{document}

\maketitle

\begin{abstract}
Event cameras record asynchronous brightness changes as streams of events, enabling efficient visual recognition at high temporal resolution. 
However, event streams often contain large numbers of redundant events, increasing unnecessary storage and computational costs. 
Dataset distillation (DD) addresses this redundancy by synthesizing a compact dataset that retains the downstream utility of the original dataset. 
DD has been widely studied on image, text, and point cloud datasets, but relatively little work has explored event datasets.
In this paper, we propose MaDeS, an event DD framework that distills raw event data by reducing both the number of samples per class and the number of events per sample. MaDeS decouples model maturity across two objectives. 
It uses randomly initialized encoders to match the feature distributions of real and synthetic events, and classifiers at different levels of maturity to strengthen class discrimination in the distilled samples.
These complementary signals help retain training utility under aggressive compression along both axes.
Extensive experiments on various datasets and distillation budgets demonstrate that MaDeS generally outperforms representative selection and distillation baselines and that the distilled data transfer well to unseen architectures.
\end{abstract}

\section{Introduction}
Event cameras report per-pixel brightness changes as an asynchronous stream of events, offering microsecond temporal resolution, high dynamic range, and low power consumption~\cite{gallego2020survey, ghosh2025survey}. These properties have motivated applications in automotive perception, robotics, and human--computer interaction~\cite{Aitsam2024appli, Gehrig2024appli}. For recognition, the stream is segmented into time windows, each forming an event sample. Common recognition approaches include dense methods~\cite{gehrig2019est}, spike methods~\cite{fang2021spike}, and point methods~\cite{ren2025secnet, ren2024eventmamba} that process samples with different representations.

\begin{figure}[t]
    \centering
    \includegraphics[width=\linewidth]{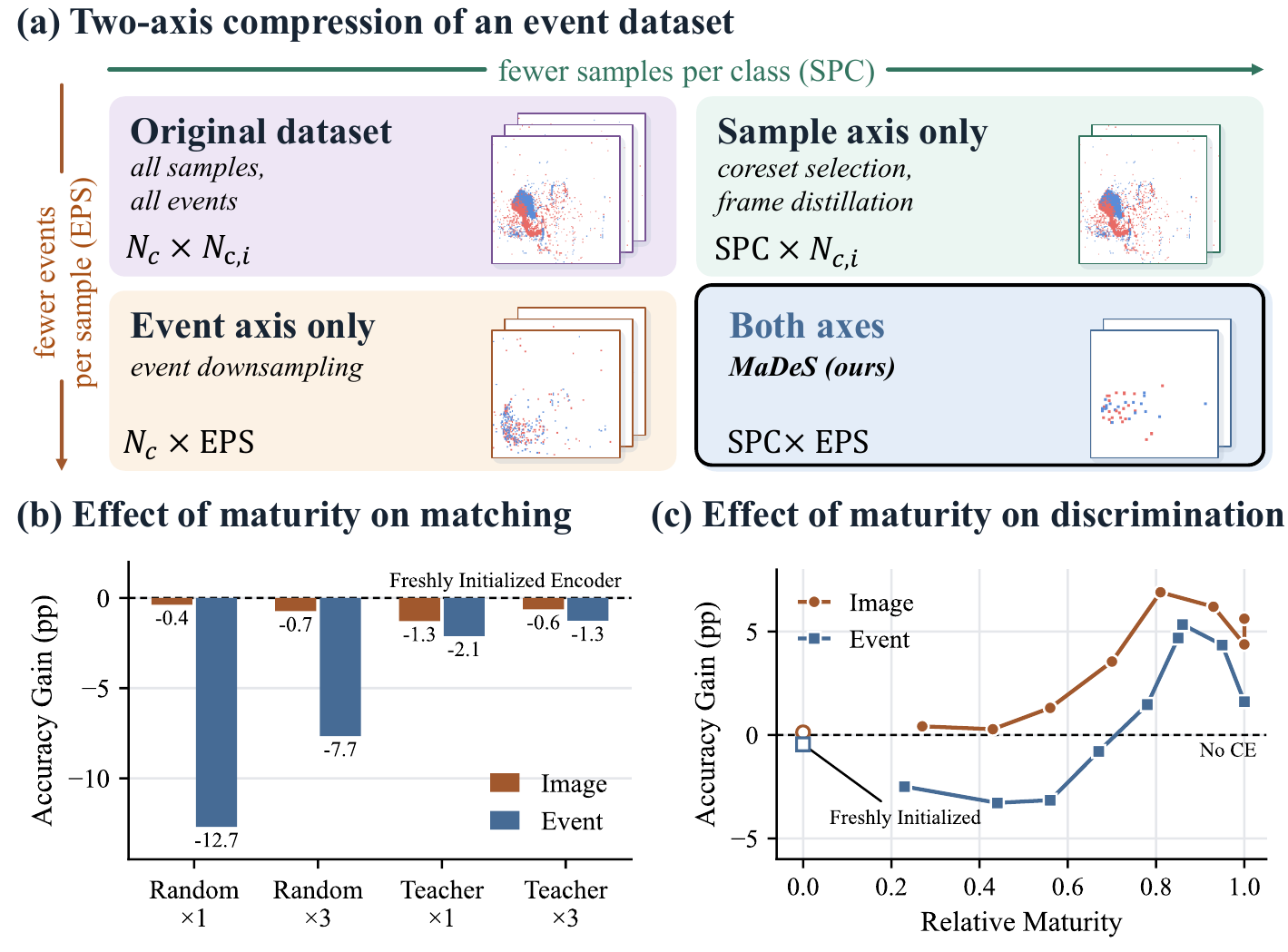}
    \caption{Two-axis event dataset distillation and the effect of model maturity. (a) Compression along different axes. (b) Accuracy gains of DM~\cite{zhao2023dm} using freshly initialized encoders, a fixed number of randomly initialized encoders, or teachers. (c) Accuracy gains of DM with cross-entropy guidance from different levels of classifier maturity.
    Analysis is conducted on CIFAR-10~\cite{krizhevsky2009cifar} for image distillation  (10 images per class) and DVS128 Gesture~\cite{amir2017gesture} for event distillation ($\mathrm{SPC} \times \mathrm{EPS} = 10 \times 128$).
    }
    \label{fig:intro}
\end{figure}

\par

Although events are spatially sparse, sensors can record millions of events per second, and datasets can span hours of recording~\cite{perot2020learning}, resulting in considerable dataset scale. Many individual events are redundant for recognition, as illustrated by the use of random event removal for data augmentation~\cite{gu2021eventdrop}. This redundancy leads to higher storage expenses and increased costs of repeatedly training models during development.

\par

Dataset distillation (DD)~\cite{wang2018dd,li2020soft,li2022awesome} reduces these costs by compressing the large dataset into a compact synthetic set that retains the downstream utility of the original data. 
It has been studied for images and point clouds~\cite{zhao2023dm,yim2025sadm}.
Temporal examples include constructing compact video sets by selecting and blending clips~\cite{ren2026protoblend} and distilling human motion sequences in the latent space of a frozen motion decoder~\cite{tian2026motion}. 
Event samples consist of asynchronous measurements with individual timestamps.
As illustrated in \cref{fig:intro}(a), an event dataset can be compressed along different axes. Reducing the number of samples per class (SPC) limits how much variation the distilled set can represent, while reducing the number of events per sample (EPS) limits the spatiotemporal detail each sample can retain. 
Reducing both yields a budget of $\text{SPC} \times \text{EPS}$ events per class. 
Moreover, downstream models map the same events to different input representations, necessitating that distilled event datasets retain useful information throughout these transformations.

\par

PACE~\cite{ye2025pace} introduces event dataset distillation for spiking neural networks by synthesizing integer event frames. However, it reduces the number of samples per class, and its frame representation cannot be used directly by point-based methods. In contrast, distribution matching~\cite{lei2023ddsurvey} offers an approach to distilling raw events by aligning features of real and synthetic samples. The original DM method~\cite{zhao2023dm} extracts features with randomly initialized models. Subsequent methods incorporate trained models into feature matching or cross-entropy (CE) guidance ~\cite{zhao2023idm,zhang2024dance}. 

\par

The choice of model depends on its role in distillation. 
In \cref{fig:intro}(b), DM with an encoder reinitialized at every step yields the best accuracy under the evaluated settings. Reusing a small number of untrained or teacher encoders incurs considerable drops on event DD compared to that on image DD. 
These comparisons suggest that, for event dataset distillation, distribution matching benefits more from diverse feature spaces than from a small set of recognition-oriented features.
The classifier used for CE guidance exhibits a different pattern, as shown in \cref{fig:intro}(c), where freshly initialized classifiers make no improvement. 
Adding CE guidance has a non-negative effect at all tested maturity levels on image DD, whereas weakly trained classifiers may decrease the accuracy on event DD. 
These observations support decoupling the model used for the two objectives based on maturity.

\par

We propose MaDeS, a framework for two-axis event dataset distillation with maturity-decoupled supervision. Specifically, MaDeS uses freshly initialized encoders to preserve diverse feature patterns through distribution matching, and classifiers selected by maturity to strengthen class discrimination through cross-entropy guidance. 
In this way, MaDeS obtains compact distilled datasets consisting of informative events that transfer across different architectures. 
We evaluate MaDeS on various datasets and distillation budgets across different architectures, showing its effectiveness and generalizability in event dataset distillation. 

\par

Our main contributions are summarized as follows:
\begin{itemize}
    \item We propose MaDeS, which distills raw event data by reducing both the number of samples per class and the number of events per sample.
    \item We decouple model maturity across two complementary objectives, using freshly initialized encoders for distribution preservation and using trained models selected by maturity to improve discrimination.
    \item We evaluate MaDeS across event benchmarks and compression budgets, demonstrating its effectiveness in event DD and generalizability across downstream architectures.
\end{itemize}

\begin{figure*}[t]
    \centering
    \includegraphics[width=17cm]{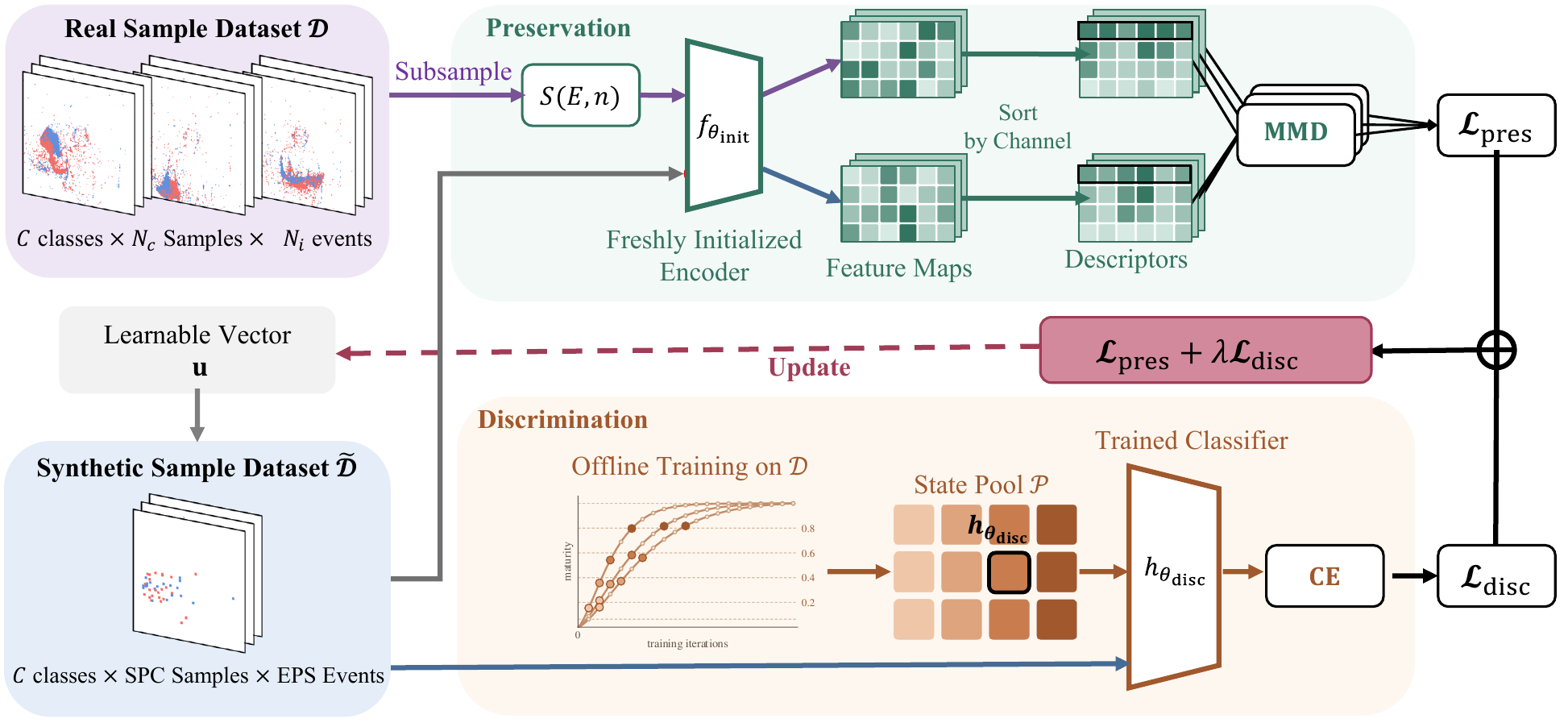}
    \caption{Overview of MaDeS. Synthetic event samples are optimized through a continuous parameterization of raw event tuples. At every iteration, the untrained encoder for distribution preservation is re-initialized, and a trained classifier for discrimination is sampled from the state pool.}
    \label{fig:workflow}
\end{figure*}

\section{Related Work}
\subsection{Event Recognition}
Event recognition assigns labels to samples segmented from the event stream, with applications including object classification~\cite{yang2025objClassi}, human action recognition~\cite{cao2026humanAct}, and fine-grained motion recognition~\cite{zhang2026fineMotion}. 
The recognition architectures differ in how they process each sample.
Dense methods convert event samples into structured representations, such as frames~\cite{gehrig2019est}, voxel grids~\cite{gehrig2019est}, or time surfaces~\cite{sironi2018hats} for conventional networks, such as convolutional neural networks~\cite{he2016resnet}.
Spike methods divide each sample into a sequence of event frames~\cite{fang2021spike}.
Recent point methods represent each sample as a spatiotemporal point cloud. 
Among them, PointNet-style architectures~\cite{qi2017pointnet,qi2017pointnet++} extract features from individual points or local neighborhoods. 
SECNet~\cite{ren2025secnet} incorporates polarity and extracts features in the frequency domain. 
EventMamba~\cite{ren2024eventmamba} models temporal dependencies across event groups.
These different processing strategies motivate distilling events that are compatible with input formats and informative across architectures.

\subsection{Dataset Distillation}
Wang~\etal~\cite{wang2018dd} introduced dataset distillation as a bilevel optimization problem to synthesize a compact dataset that retains the training utility of the original dataset.
Subsequent methods distill datasets by matching gradients~\cite{zhao2021dc,zhao2021dsa}, training trajectories~\cite{cazenavette2022mtt}, or feature distributions~\cite{zhao2023dm}.
Parameter-matching methods can discard parameters that are difficult to match~\cite{li2024pruning} or learn importance weights for individual parameters~\cite{li2024iadd}.

\par

Among distribution matching methods, DM~\cite{zhao2023dm} aligns class-wise feature means using randomly initialized encoders.
Later work enriches the matched statistics~\cite{zhang2024m3d,wang2025ncfm, yin2023sre2l} and matching strategies ~\cite{wang2022cafe}.
Hyperbolic dataset distillation matches feature distributions in a space that represents hierarchy~\cite{li2025hyperbolic}, while difficulty-aware matching computes separate running statistics for groups of examples~\cite{ma2026difficulty}.
IDM~\cite{zhao2023idm} shares models from an online pool between feature matching and CE regularization. 
DANCE~\cite{zhang2024dance} uses trained experts for CE-based class calibration, and interpolates initial and trained weights for feature matching.
WMDD~\cite{liu2025wmdd} matches Wasserstein barycenters of features from a trained teacher. 
For frozen vision encoders, CLP-DD differentiates through a closed-form linear probe~\cite{peng2026clpdd}, and SRG uses self-supervised features to guide diffusion toward class prototypes~\cite{li2026srg}. Other forms of class supervision include soft labels for distilled medical images~\cite{li2022soft} and prototype constraints for fine-grained recognition~\cite{ma2026fd2}.
Distribution matching has also been applied to point cloud distillation~\cite{bo2025point}. SADM~\cite{yim2025sadm} aligns semantically corresponding points within each channel before matching and jointly optimizes the orientation of each synthetic cloud. 
However, the application of distribution matching to asynchronous and temporal event datasets remains underexplored.

\par
The budget within a sample also matters. In image distillation, methods preserve detail by matching global and local features~\cite{li2024global}, choosing crops according to foreground occupancy~\cite{li2026foreground}, or combining discriminative patches from the same class~\cite{fan2026deco}. Selection across generated images can use estimated classification difficulty~\cite{li2025sampling} or semantic relevance and diversity~\cite{li2026sas}. For event streams, the two corresponding choices are which recordings to represent and which timestamped events to retain within each sample.

\subsection{Data-efficient Learning for Event Data}
Point-based event networks typically subsample events before feature extraction~\cite{ren2025secnet, ren2024eventmamba}. 
Farthest point sampling (FPS), used in PointNet++~\cite{qi2017pointnet++}, selects points that progressively expand spatial coverage.
Event-specific downsampling methods use criteria such as local event density~\cite{araghi2025density} or estimates of event importance~\cite{girbau2025poled}.
For dataset distillation, PACE~\cite{ye2025pace} synthesizes dense integer event frames for spike methods by matching spatial and temporal statistics of features derived from membrane potentials. However, aggregating events into frames loses information needed to recover individual events, which limits the use of the distilled dataset with point-based architectures.

\section{Method}
MaDeS learns a compact dataset directly in the raw event format with reduced SPC and EPS. As shown in \cref{fig:workflow}, it uses randomly initialized encoders to preserve the distribution of real events and trained classifiers of varying maturity to retain discriminative cues between classes.

\subsection{Problem Formulation}
An event camera records a stream of events $(x,y,t,p)$, where $(x,y)$ is the pixel location, $t$ is the timestamp, and $p\in\{-1,+1\}$ is the polarity of the brightness change.
For recognition, the stream is segmented into time windows, and each window is treated as one labeled sample. 
Let $\mathcal{D}=\{(E_{c,i},c)\}_{1\le c \le C, \ 1\le i\le N_c}$
denote the real dataset, where $C$ is the number of classes,
$N_c$ is the number of samples in class $c$, and $E_{c,i}$ contains
$N_{c,i}$ events.

Given a budget of SPC samples per class and EPS events per sample,
event dataset distillation learns a synthetic set
$\tilde{\mathcal{D}}=\{(\tilde{E}_{c,i},c)\}_{1\le c\le C, \ 1\le i\le \mathrm{SPC}}$,
where each $\tilde{E}_{c,i}$ contains exactly $\mathrm{EPS}$ events.
Typically, $\mathrm{SPC} \ll N_c$ and $\mathrm{EPS} \ll N_{c,i}$.
The goal is for models trained from scratch on $\tilde{\mathcal{D}}$ to achieve downstream performance comparable to that of models trained on the real dataset $\mathcal{D}$, both for the architecture used in distillation and for unseen architectures.

\subsection{Data Preparation}
\label{sec:param}
For each real event sample, we normalize the spatial coordinates and timestamps to $[0,1]$ for range consistency. 
We match real and synthetic samples at the same EPS to keep their feature density with a subsample operator $\mathcal{S}(E, n)$ that randomly draws $n$ events uniformly from the sample $E$ while preserving their temporal order. The synthetic set is initialized by randomly selecting SPC real samples per class and applying $\mathcal{S}$ to obtain EPS events per sample.

\par

To optimize synthetic events by gradient descent, we relax their attributes to continuous values and parameterize each event with an unconstrained vector $\mathbf{u}=(u_x,u_y,u_t,u_p)$ mapped to the normalized event domain as follows: 
\begin{equation}
\begin{aligned}
  \tilde{x} = \sigma(u_x), \quad \tilde{y}=\sigma(u_y), \\ \tilde{t}=\sigma(u_t), \quad \tilde{p}=\tanh(u_p),
\end{aligned}
\end{equation}
where $\sigma$ is the sigmoid function. 
We optimize $\mathbf{u}$ and recompute the event attributes from it after each update.
The smooth mappings keep the attributes within their prescribed ranges and avoid zero-gradient flat regions~\cite{carlini2027repara}.

\subsection{Distribution Preservation}
Distribution matching embeds real and synthetic data in a shared feature space and aligns their class-wise feature distributions~\cite{zhao2023dm}. 
Motivated by the advantage of reinitialization in \cref{fig:intro}(b),
we assign fresh random weights to the shared encoder $f_{\theta_{\mathrm{init}}}$ at each distillation step. 
To represent the real distribution, for each class $c$, we draw a real batch $\mathcal{B}_c= \{ E_{c,b} \}_{b=1}^{B}$  from $\mathcal{D}$ and subsample each real sample to the target EPS as follows:
\begin{equation}
    \mathcal{B}_c^* = \{ E_{c,b}^* \}_{b=1}^{B} = \{ \mathcal{S}(E_{c,b}, \text{EPS}) \}_{b=1}^{B}.
\end{equation}

\par

Equal event counts make the features comparable in size, but events at the same index can occur at different times across samples. Some point-based encoders even further reorder them, making the same row in the feature map refer to an arbitrary event or group, with no correspondence across samples. Directly aligning rows therefore compares unrelated local features. Global pooling gives a fixed-length descriptor but can obscure variation among local responses, which are important when compressing events within each sample. Inspired by SADM~\cite{yim2025sadm}, we instead align local responses by rank within each channel.

\par

A hierarchical point-based encoder such as SECNet~\cite{ren2025secnet} progressively subsamples and aggregates events, producing feature maps with varying spatial resolutions at different stages.
We match features separately at all $S$ stages to capture fine-grained and aggregated event structure. 
At stage $s$, the feature map for $E_{c,b}^{*}$ has $n_s$ rows representing points or local groups and $d_s$ feature channels. We sort the responses in descending order within each channel and flatten the resulting map to obtain a descriptor $\mathbf{z}^s_{c,b}$ as follows:
\begin{equation}
    \mathbf{z}^s_{c,b} = \operatorname{vec}\bigl( \operatorname{sort}_{\downarrow} f^s_{\theta_{\mathrm{init}}}(E_{c,b}^*) \bigr), 
\end{equation}
where $\operatorname{sort}_{\downarrow}$ sorts each channel along the point dimension, $\operatorname{vec}$ flattens the sorted map, and $f^s_{\theta_{\mathrm{init}}}$ denotes the output of the encoder parameterized by $\theta_{\mathrm{init}}$ at stage $s$. 
The encoder incorporates temporal information into responses, and sorting makes the descriptor invariant to permutations of the feature-map rows. 
Following SADM, we also collect the maximum response in each channel to form the top-1 descriptor $\mathbf{m}_{c,b}^s$ as follows:
\begin{equation}
    \mathbf{m}^s_{c,b} = \max_{1\le a\le n_s} f^s_{\theta_{\mathrm{init}}}(E_{c,b}^*)_{a,:}, 
\end{equation}
where the maximum is taken separately for each channel. We compute both descriptors in the same way for synthetic samples.

\par

At each stage $s$, let
$\mathcal{Z}_c^s=\{\mathbf{z}^s_{c,b}\}_{b=1}^{B}$ and $\tilde{\mathcal{Z}}^s_c=\{ \tilde{\mathbf{z}}^s_{c, i} \}_{i=1}^{\text{SPC}}$ denote the collections of full descriptors for the real and synthetic samples of class $c$. 
Let $\mathcal{M}_c^s=\{\mathbf{m}^s_{c,b}\}_{b=1}^{B}$ and $\tilde{\mathcal{M}}_c^s = \{ \tilde{\mathbf{m}}^s_{c,i} \}_{i=1}^{\text{SPC}}$ denote the corresponding top-1 descriptor collections. The preservation objective sums the discrepancies over all classes and stages as follows:
\begin{equation}
\begin{aligned}
\mathcal{L}_{\mathrm{pres}} = \sum_{c=1}^{C} \sum_{s=1}^{S} \bigl[ \operatorname{MMD}(\mathcal{Z}_c^s,\tilde{\mathcal{Z}}_c^s) \\ + \gamma \ \operatorname{MMD} (\mathcal{M}_c^s,\tilde{\mathcal{M}}_c^s) \bigr],
\end{aligned}
\label{eq}
\end{equation}
where $\operatorname{MMD}$ denotes the squared maximum mean discrepancy computed with a multi-bandwidth Gaussian kernel and $\gamma$ weights the top-1 term. 

\begin{table*}[t]
\centering
\footnotesize
\setlength{\tabcolsep}{4pt}
\caption{Comparison of test accuracy (\%) between MaDeS and baseline methods on DVS128 Gesture and N-MNIST under different SPC and EPS settings. ``Full Dataset'' denotes all training samples, each subsampled to the target EPS. 
The highest and second-highest means are shown in bold and underlined, respectively.}
\label{tab:main}
\begin{tabular}{l|ccccc|ccccc}
\toprule

& \multicolumn{5}{c|}{DVS128 Gesture~\cite{amir2017gesture}}
& \multicolumn{5}{c}{N-MNIST~\cite{orchard2015nmnist}} 
\\

SPC$\times$EPS
& $1 \times 128$ & $10 \times 128$ 
& $1 \times 256$ & $10 \times 256$
& $10 \times 512$
& $1 \times 128$ & $10 \times 128$ 
& $1 \times 256$ & $10 \times 256$
& $10 \times 512$
\\

{\scriptsize Compression Ratio}
& \scriptsize{21k$\times$} 
& \scriptsize{2.1k$\times$} 
& \scriptsize{11k$\times$} 
& \scriptsize{1.1k$\times$}
& \scriptsize{0.5k$\times$}
& \scriptsize{192k$\times$} 
& \scriptsize{19k$\times$} 
& \scriptsize{96k$\times$}
& \scriptsize{9.6k$\times$}
& \scriptsize{4.8k$\times$}
\\
\midrule

Random
& $30.1_{\pm3.0}$ & $58.0_{\pm1.4}$
& $26.5_{\pm3.6}$ & $61.7_{\pm1.8}$
& $57.6_{\pm2.3}$
& $25.8_{\pm2.7}$ & $61.2_{\pm2.1}$
& $31.6_{\pm3.1}$ & $78.1_{\pm0.7}$
& $85.1_{\pm0.7}$
\\

Herding~\cite{welling2009herding}
& $33.7_{\pm4.4}$ & $61.7_{\pm0.8}$
& $31.0_{\pm3.7}$ & $64.6_{\pm2.3}$
& $57.8_{\pm3.8}$
& $28.2_{\pm2.9}$ & $63.2_{\pm1.3}$
& $40.3_{\pm6.3}$ & $81.8_{\pm1.3}$
& $88.7_{\pm1.0}$
\\

Closest
& $33.4_{\pm4.4}$ & $57.1_{\pm2.2}$
& $31.0_{\pm3.1}$ & $58.8_{\pm3.1}$
& $57.4_{\pm2.5}$
& $29.0_{\pm2.4}$ & $64.7_{\pm2.4}$
& $39.7_{\pm5.7}$ & $79.7_{\pm1.4}$
& $86.3_{\pm1.1}$
\\

K-Center~\cite{sener2018kcenter}
& $34.0_{\pm3.7}$ & $48.4_{\pm2.9}$
& $31.1_{\pm3.1}$ & $47.0_{\pm2.3}$
& $45.8_{\pm2.6}$
& $27.9_{\pm2.5}$ & $40.5_{\pm2.5}$
& $40.2_{\pm5.8}$ & $54.7_{\pm3.5}$
& $65.2_{\pm4.0}$
\\

Stride
& $33.8_{\pm2.8}$ & $58.0_{\pm2.3}$
& $31.9_{\pm3.2}$ & $61.4_{\pm3.2}$
& $57.9_{\pm5.0}$
& $26.4_{\pm5.1}$ & $61.1_{\pm1.6}$
& $33.0_{\pm2.8}$ & $77.2_{\pm2.0}$
& $86.2_{\pm1.3}$
\\

FPS~\cite{qi2017pointnet++}
& $18.7_{\pm2.5}$ & $30.5_{\pm3.6}$
& $22.1_{\pm2.4}$ & $37.7_{\pm3.5}$
& $49.2_{\pm4.4}$
& $12.6_{\pm1.5}$ & $23.5_{\pm3.5}$
& $18.2_{\pm4.3}$ & $55.6_{\pm2.9}$
& $70.3_{\pm2.7}$
\\

Araghi \etal~\cite{araghi2025density}
& $26.7_{\pm3.0}$ & $57.4_{\pm1.8}$
& $26.5_{\pm3.2}$ & $59.1_{\pm3.5}$
& $55.1_{\pm4.3}$
& $27.1_{\pm5.8}$ & $58.8_{\pm2.2}$
& $33.4_{\pm2.4}$ & $73.8_{\pm1.5}$
& $81.4_{\pm1.9}$
\\

POLED~\cite{girbau2025poled}
& $27.3_{\pm3.6}$ & $55.3_{\pm2.9}$
& $27.5_{\pm2.3}$ & $59.6_{\pm2.9}$
& $55.4_{\pm3.5}$
& $28.7_{\pm5.9}$ & $60.2_{\pm2.2}$
& $31.5_{\pm5.8}$ & $73.2_{\pm1.6}$
& $81.9_{\pm2.6}$
\\

DM~\cite{zhao2023dm}
& $31.6_{\pm3.3}$ & $64.9_{\pm1.0}$
& $30.3_{\pm2.1}$ & $66.6_{\pm1.8}$
& $61.5_{\pm4.4}$
& $40.3_{\pm2.3}$ & $70.1_{\pm0.9}$
& $43.8_{\pm3.6}$ & $82.4_{\pm0.8}$
& $86.3_{\pm1.2}$
\\

IDM~\cite{zhao2023idm}
& \bm{$41.1_{\pm2.6}$} & $\underline{69.7}_{\pm1.1}$
& \bm{$39.1_{\pm3.5}$} & $71.7_{\pm2.7}$
& $66.6_{\pm5.1}$
& \bm{$46.0_{\pm3.9}$} & $69.4_{\pm1.3}$
& $54.8_{\pm3.6}$ & $80.9_{\pm0.9}$
& $87.2_{\pm1.0}$
\\

SADM~\cite{yim2025sadm}
& $33.5_{\pm3.2}$ & $68.0_{\pm0.9}$
& $34.0_{\pm2.3}$ & $\underline{71.8}_{\pm2.8}$
& $\underline{66.7}_{\pm3.1}$
& $\underline{45.7}_{\pm2.6}$ & $\underline{72.2}_{\pm0.8}$
& $\underline{55.0}_{\pm4.2}$ & $\underline{85.2}_{\pm0.7}$
& $\underline{90.2}_{\pm0.9}$
\\

WMDD~\cite{liu2025wmdd}
& $31.1_{\pm2.7}$ & $57.2_{\pm1.2}$
& $29.9_{\pm2.8}$ & $59.8_{\pm2.0}$
& $55.6_{\pm3.4}$
& $29.6_{\pm3.9}$ & $62.6_{\pm1.4}$
& $33.5_{\pm3.9}$ & $77.2_{\pm1.3}$
& $84.9_{\pm0.5}$
\\

\rowcolor{MaDeSRow}
MaDeS
& $\underline{37.1}_{\pm4.6}$ & \bm{$73.4_{\pm1.1}$}
& $\underline{36.5}_{\pm3.2}$ & \bm{$78.5_{\pm1.7}$}
& \bm{$77.7_{\pm3.0}$}
& $45.5_{\pm2.5}$ & \bm{$76.9_{\pm1.0}$}
& \bm{$57.2_{\pm4.1}$} & \bm{$87.2_{\pm0.5}$}
& \bm{$91.4_{\pm0.5}$}
\\

\midrule
Full Dataset
& \multicolumn{2}{c}{$90.1_{\pm0.2}$}
& \multicolumn{2}{c}{$94.1_{\pm0.1}$}
& \multicolumn{1}{c|}{$95.5_{\pm0.2}$}
& \multicolumn{2}{c}{$92.4_{\pm0.1}$}
& \multicolumn{2}{c}{$96.4_{\pm0.1}$}
& \multicolumn{1}{c}{$98.4_{\pm0.1}$}
\\
\bottomrule

\end{tabular}
\end{table*}

\subsection{Inter-Class Discrimination}
Distribution matching aligns real and synthetic features within each class. However, at low EPS, subsampling can discard distinguishing cues and leave samples from different classes with similar features. We add a cross-entropy (CE) objective that uses a classifier's logits to guide the synthetic events toward patterns that separate the assigned class from competing classes.

\par
\cref{fig:intro}(c) shows that weakly trained classifiers can reduce downstream accuracy on the event dataset, and that the gains vary among trained checkpoints. 
Models tend to learn simpler patterns early in training and capture more difficult ones as training progresses~\cite{guo2024datm}. As a result, checkpoints at different levels of maturity place different emphasis on class-discriminative cues. We therefore draw classifiers from multiple training runs and maturity levels to offer supervision from different decision functions.
\par

Specifically, we measure maturity by accuracy on real samples. We train $R$ models on the real training set at the target EPS and save $K$ checkpoints per run.
Let $s_{r,j}$ denote the maturity of checkpoint $j$ in run $r$, and let $s_r^\mathrm{max}$ and $s_r^\mathrm{min}$ be the maximum and minimum maturity values among the checkpoints in that run. Given a target fraction $q\in(0,1)$, the checkpoint closest to the target maturity is selected as follows:
\begin{equation}
j_{r,q} = \arg\min_{1\le j\le K}  \bigl| s_r^\mathrm{min} + q (s_r^\mathrm{max}-s_r^\mathrm{min}) - s_{r,j}  \bigr|.
\end{equation}
The selected checkpoints across all runs and target fractions form the state pool $\mathcal{P}$.


\par

At each distillation step, we sample a checkpoint $\theta_{\mathrm{disc}}$ uniformly from $\mathcal P$. The inter-class discrimination loss is obtained from its logits on the synthetic samples $\tilde{E}_{c,i}$ as follows:
\begin{equation}
  \mathcal{L}_{\mathrm{disc}} =  \frac{1}{C \times \text{SPC}} \sum_{c=1}^{C} \sum_{i=1}^{\text{SPC}} \operatorname{CE} \bigl( h_{\theta_{\mathrm{disc}}}(\tilde{E}_{c,i}), c \bigr),
  \label{eq:disc}
\end{equation}
where $h_{\theta_{\mathrm{disc}}}$ is the classifier parameterized by $\theta_{\mathrm{disc}}$ and $\operatorname{CE}$ denotes the cross-entropy function.

\subsection{Overall Objective}
\label{sec:overall}

We optimize the synthetic event samples using the combined preservation loss $\mathcal{L}_{\mathrm{pres}}$ and discrimination loss $\mathcal{L}_{\mathrm{disc}}$ as follows
\begin{equation}
 \mathcal{L}_{\mathrm{MaDeS}} = \mathcal{L}_{\mathrm{pres}} + \lambda \ \mathcal{L}_{\mathrm{disc}},
  \label{eq:total}
\end{equation}
where $\lambda$ is the weight for the discrimination term.
During distillation, we optimize the parameter vectors $\mathbf{u}$ through the event reparameterization and sort the mapped events by timestamp before passing them to the encoder and classifier. 
MaDeS aligns model maturity with each distillation objective, using freshly initialized encoders to preserve diverse spatiotemporal structure through feature matching and trained classifiers of varying maturity to emphasize cues that distinguish classes within the limited event budget.
The raw event tuples are stored as the distilled dataset that can be used across different architectures.

\section{Experiments}
\label{sec:exp}

\subsection{Experimental Settings}
\label{sec:setup}
We evaluate on four event recognition benchmarks from two sources.
Natively recorded datasets capture real scenes with an event camera.
DVS128 Gesture~\cite{amir2017gesture} contains hand and arm gestures from different people and lighting conditions. We use 10 classes after removing the \emph{other} class. 
THU-EACT-50~\cite{gao2023thu1, gao2024thu2} contains 50 human action classes performed by different people.
Converted datasets are neuromorphic versions of static image benchmarks, recorded from images displayed on a monitor while the sensor performs three saccade-like motions.
N-MNIST~\cite{orchard2015nmnist} and N-Caltech101~\cite{orchard2015nmnist} contain event samples for 10 digit classes and 101 object classes, respectively.

\par

We use SECNet~\cite{ren2025secnet} as the default distillation model, which is a point-based network.
For cross-architecture evaluation, we include EventMamba (EM)~\cite{ren2024eventmamba}, a standard ResNet-18~\cite{he2016resnet} that takes binned event grids as input (Voxel), and a spiking VGG network (VGG-SNN)~\cite{sengupta2019vgg}, covering three common architecture families.
We construct the event samples following the SECNet protocol and compute all compression ratios relative to the sample dataset.

\par

For baselines, we follow the published baseline settings, except that per-object rotation in SADM is omitted to accommodate the event cloud, and soft-label relabeling in WMDD is excluded to ensure a fair evaluation. 
For MaDeS, we set $\gamma = 0.5$ for $\mathcal{L}_\mathrm{pres}$ and use dataset-specific values of $\lambda$ to account for differences in classification difficulty. 
Specifically, we set $\lambda=0.01$ for DVS128 Gesture and THU-EACT-50, $\lambda=0.003$ for N-MNIST, and $\lambda=0.1$ for N-Caltech101.
The real batch size differs across datasets and distillation budgets. 
To construct the state pool, we save $K=20$ checkpoints for each of $R=3$ independent training runs and use a set of target fractions $q \in \{0.2,0.4,0.6,0.8\}$, forming a state pool of 12 checkpoints. 
We build the pool once per dataset and EPS. 

\par

For evaluation, we train a randomly initialized network from scratch for 3{,}000 iterations and evaluate it on real test windows. 
We report window-level accuracy over three distillations, each with three evaluation runs.
All experiments are conducted on NVIDIA H100 GPUs.
More detailed settings are listed in the supplementary material.

\subsection{Benchmark Results}
We compare MaDeS with three families of baselines on DVS128 Gesture and N-MNIST. Sample selection methods include random selection (Random), selection of samples closest to the prototypes (Closest), Herding~\cite{welling2009herding}, and K-Center~\cite{sener2018kcenter}, using the subsample operator $\mathcal{S}$ to reduce EPS.
Event selection methods include subsampling at a fixed stride (Stride), FPS~\cite{qi2017pointnet++}, density-based selection by Araghi \etal~\cite{araghi2025density}, and POLED~\cite{girbau2025poled}, using random selection to reduce SPC.
Dataset distillation baselines include DM~\cite{zhao2023dm}, IDM~\cite{zhao2023idm}, SADM~\cite{yim2025sadm}, and WMDD~\cite{liu2025wmdd}.
We also provide a reference using all training samples subsampled to the target EPS (Full Dataset). 
\par

As shown in \cref{tab:main}, MaDeS achieves the best performance among the compared selection and distillation methods in most settings. IDM leads at SPC $=1$ on DVS128 Gesture and at $\text{SPC} \times \text{EPS} = 1 \times 128$ on N-MNIST, but it becomes less competitive as the distillation budget increases.
On DVS128 Gesture, accuracy decreases for most methods when the budget increases from $1\times128$ to $1\times256$ or from $10\times256$ to $10\times512$. 
We examine this trend across a wider range of budgets in \cref{sec:scale} and attribute it to overfitting on sample-specific detail at low SPC.
Overall, MaDeS remains effective and stable across different settings, showing the effectiveness of maturity-decoupled objectives: freshly initialized encoders offer diverse feature patterns for preservation, and the trained classifiers with varying maturity cover different decision functions for discrimination.

\begin{table}[t]
\centering
\footnotesize
\setlength{\tabcolsep}{5pt}
\caption{Test accuracy (\%) under different $\mathrm{SPC}\times\mathrm{EPS}$ budgets. The highest and second-highest means are shown in bold and underlined, respectively.}
\label{tab:scale}
\begin{tabular}{lcccc}
\toprule
\textbf{Method} & $1\times128$ & $10\times128$ & $1\times256$ & $10\times256$ \\
\midrule
\multicolumn{5}{l}{\textbf{THU-EACT-50} (50 classes)} \\[2pt]
Random
& $5.1_{\pm0.4}$ & $13.9_{\pm1.3}$
& $6.1_{\pm0.6}$ & $19.0_{\pm1.4}$
\\

Herding~\cite{welling2009herding}
& $6.8_{\pm1.0}$ & $15.2_{\pm1.4}$
& $7.4_{\pm0.6}$ & $20.3_{\pm1.9}$
\\

DM~\cite{zhao2023dm}
& $10.6_{\pm1.2}$ & $15.6_{\pm1.3}$
&$8.6_{\pm0.7}$ & $21.3_{\pm1.4}$
\\

IDM~\cite{zhao2023idm}
& $8.7_{\pm0.8}$ & $\underline{23.7}_{\pm1.8}$
& $8.7_{\pm1.0}$ & $\underline{30.0}_{\pm1.3}$
\\

SADM~\cite{yim2025sadm}
& $\underline{12.1}_{\pm1.4}$ & $17.3_{\pm1.8}$
& $\underline{11.8}_{\pm1.1}$ & $25.2_{\pm1.2}$
\\

\rowcolor{MaDeSRow}
MaDeS
& \bm{$13.6_{\pm2.3}$} & \bm{$25.8_{\pm1.5}$}
& \bm{$13.7_{\pm1.9}$} & \bm{$39.6_{\pm2.2}$}
\\
\midrule
\multicolumn{5}{l}{\textbf{N-Caltech101} (101 classes)} \\[2pt]
Random
& $5.4_{\pm1.7}$ & $13.6_{\pm0.5}$
& $7.5_{\pm2.2}$ & $18.6_{\pm1.3}$
\\

Herding~\cite{welling2009herding}
& $5.9_{\pm0.7}$ & $14.6_{\pm0.4}$
& $8.9_{\pm1.5}$ & $21.1_{\pm0.4}$
\\

DM~\cite{zhao2023dm}
& $8.9_{\pm1.7}$ & $16.9_{\pm0.6}$
& $9.4_{\pm1.9}$ & $23.2_{\pm0.7}$
\\

IDM~\cite{zhao2023idm}
& $8.2_{\pm1.9}$ & $\underline{22.8}_{\pm0.7}$
& $10.0_{\pm1.9}$ & $\underline{31.0}_{\pm0.7}$
\\

SADM~\cite{yim2025sadm}
& $\underline{9.5}_{\pm1.6}$ & $17.8_{\pm1.0}$
& $\underline{11.9}_{\pm1.0}$ & $24.8_{\pm0.9}$
\\

\rowcolor{MaDeSRow}
MaDeS
& \bm{$13.3_{\pm0.9}$} & \bm{$26.9_{\pm0.9}$}
& \bm{$16.2_{\pm0.9}$} & \bm{$34.4_{\pm0.9}$}
\\
\bottomrule
\end{tabular}
\end{table}

\subsection{Scalability to Datasets with More Classes}
We also compare MaDeS with representative baselines across distillation budgets on the THU-EACT-50 and N-Caltech101 datasets,  which have larger label spaces. As shown in \cref{tab:scale}, MaDeS achieves the highest accuracy across the evaluated distillation budgets on both datasets.
We also find that IDM is less competitive at SPC $=1$ on these datasets than on DVS128 Gesture. 
This may stem from the limited maturity of the classifiers in IDM's online pool. 
IDM maintains an online pool by adding randomly initialized models and periodically training a subset of them during distillation, resulting in a limited number of updates per model. 
On 10-class DVS128 Gesture, classifiers reach sufficient maturity to provide useful class supervision after relatively few updates.
On datasets with more classes, classifiers require more updates to reach that level of maturity, and models from the online pool provide weaker supervision.
Instead, MaDeS selects offline checkpoints of a range of maturity values based on their observed maturity, accommodating differences in learning speed across datasets, which also avoids repeated online training. 
We report additional validation in the supplementary material.

\subsection{Cross-Architecture Generalization}
\label{sec:cross}
\begin{table}[t]
\centering
\footnotesize
\setlength{\tabcolsep}{3pt}
\caption{Cross-architecture evaluation on DVS128 Gesture using SECNet, EventMamba (EM), ResNet-18 (Voxel), and VGG-SNN under $\text{SPC} \times \text{EPS} = 10 \times 128$. The datasets are distilled with SECNet. The highest and second-highest means are shown in bold and underlined, respectively.}
\label{tab:cross}
\begin{tabular}{l|cccc}
\toprule

\multirow{2}{*}{Distillation} 
& \multicolumn{4}{c}{Evaluation} 
\\
& SECNet~\cite{ren2025secnet} 
& EM~\cite{ren2024eventmamba} 
& Voxel~\cite{he2016resnet}
& VGG-SNN~\cite{sengupta2019vgg} 
\\
\midrule

Random 
& $58.0_{\pm1.4}$ & $43.5_{\pm1.1}$ 
& $56.9_{\pm4.1}$ & $66.7_{\pm2.4}$ 
\\

Herding~\cite{welling2009herding}
& $61.7_{\pm0.8}$ & $44.6_{\pm2.4}$ 
& $56.8_{\pm1.4}$ & $69.0_{\pm1.0}$ 
\\

DM~\cite{zhao2023dm} 
& $64.9_{\pm1.0}$ & $45.2_{\pm1.4}$ 
& $55.5_{\pm3.9}$ & $70.6_{\pm2.8}$ 
\\

IDM~\cite{zhao2023idm} 
& $\underline{69.7}_{\pm1.1}$ & $\underline{52.4}_{\pm0.5}$ 
& $\underline{60.9}_{\pm2.5}$ & $\underline{76.8}_{\pm0.6}$ 
\\

SADM~\cite{yim2025sadm} 
& $68.0_{\pm0.9}$ & $47.7_{\pm2.0}$ 
& $58.6_{\pm1.9}$ & $72.6_{\pm1.6}$ 
\\

\rowcolor{MaDeSRow}
MaDeS 
& \bm{$73.4_{\pm1.1}$} & \bm{$53.7_{\pm2.5}$} 
& \bm{$61.2_{\pm0.9}$} & \bm{$78.3_{\pm1.3}$} 
\\

\bottomrule
\end{tabular}
\end{table}

We validate the cross-architecture generalizability of MaDeS by evaluating the SECNet-distilled datasets on unseen architectures on DVS128 Gesture under the distillation budget of $\text{SPC} \times \text{EPS} = 10 \times 128$.
As shown in \cref{tab:cross}, MaDeS achieves the highest mean accuracy with SECNet and with all unseen architectures. 
Different methods show a similar trend on different evaluation architectures. 
IDM also outperforms SADM across the evaluated architectures, indicating that class-discriminative supervision is more useful for event dataset distillation, possibly due to the decreased EPS that affects the reliability of distribution matching.
We examine the contributions of preservation and discrimination directly in \cref{sec:ablation}.
We also include comparisons on other datasets and using other distillation architectures in the supplementary material.

\begin{figure*}[t]
    \centering
    \includegraphics[width=17cm]{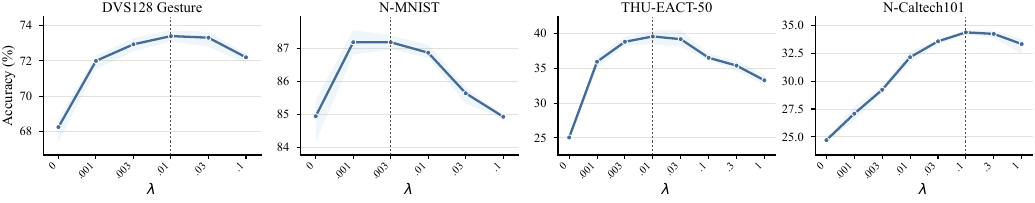}
    \caption{Sensitivity of downstream accuracy to the discrimination weight $\lambda$ across datasets. Shaded regions indicate the minimum-to-maximum range. Dashed lines mark the values used in the main experiments.}
    \label{fig:lambda}
\end{figure*}

\begin{table}[t]
\centering
\footnotesize
\setlength{\tabcolsep}{5pt}
\caption{Accuracy (\%) of MaDeS on DVS128 Gesture over distillation budgets. Values in parentheses are differences from Herding in percentage points. Darker shading indicates larger gains.}
\label{tab:ipc}
\begin{tabular}{c|cccc}
\toprule
SPC $\backslash$ EPS & 128 & 256 & 512 & 1024 \\
\midrule
1
& \gaincell{37.1}{+3.4}
& \gaincell{36.5}{+5.5}
& \gaincell{34.6}{+3.4}
& \gaincell{33.1}{+6.2}
\\

5
& \gaincell{67.2}{+13.8}
& \gaincell{67.5}{+15.3}
& \gaincell{62.4}{+15.5}
& \gaincell{60.3}{+17.2}
\\

10
& \gaincell{73.4}{+11.7}
& \gaincell{78.5}{+13.9}
& \gaincell{77.7}{+19.9}
& \gaincell{77.6}{+22.3}
\\

20
& \gaincell{76.8}{+7.3}
& \gaincell{81.8}{+5.7}
& \gaincell{83.7}{+10.9}
& \gaincell{83.2}{+10.6}
\\

50
& \gaincell{78.8}{+3.5}
& \gaincell{83.8}{+1.4}
& \gaincell{85.9}{+2.1}
& \gaincell{84.5}{+0.7}
\\

\bottomrule
\end{tabular}
\end{table}

\subsection{Two-Axis Compression}
\label{sec:scale}
We further evaluate MaDeS under more distillation budgets with $\text{SPC} \in \{ 1,5,10,20,50 \}$ and $\text{EPS} \in \{ 128, 256,512,1024 \}$ on DVS128 Gesture. We also include the comparison with Herding. As shown in \cref{tab:ipc}, MaDeS outperforms Herding throughout the grid, while the gains vary across budgets.
Increasing SPC improves accuracy at each EPS, with diminishing improvements at the largest budgets.
Increasing EPS sometimes results in reduced accuracy. 
The advantage over Herding also narrows at the largest SPC, suggesting that a larger subset of real samples covers enough recognition information.
We also notice that, at low SPC, fewer events per sample yield higher accuracy. One possible explanation is that denser event samples retain more sample-specific details, which models may overfit to when trained on only a few such samples.
The disadvantage recedes as more samples become available to cover within-class variation. We provide the evaluation on other datasets in the supplementary material.

\subsection{Ablation on Distillation Objective}
\label{sec:ablation}
\begin{table}[t]
\centering
\footnotesize
\setlength{\tabcolsep}{4pt}
\caption{Ablation of the loss terms on DVS128 Gesture under $\text{SPC} \times \text{EPS} = 10 \times 128$. The datasets are distilled with SECNet. $\bigcirc$ denotes replacing $\mathcal{L}_{\mathrm{pres}}$ with standard matching of DM~\cite{zhao2023dm}. The best values are highlighted in bold.}
\label{tab:ablation}
\begin{tabular}{ll|cccc}
\toprule

$\mathcal{L}_\mathrm{pres}$ & $\mathcal{L}_\mathrm{disc}$
& SECNet & EM & Voxel & VGG-SNN 
\\
\midrule
$\bigcirc$ & $\times$
& $64.9_{\pm1.0}$ & $45.2_{\pm1.4}$ 
& $55.5_{\pm3.9}$ & $70.6_{\pm2.8}$
\\

$\times$ & $\checkmark$
& $66.2_{\pm1.3}$ & $47.7_{\pm1.8}$
& $59.0_{\pm4.0}$ & $70.0_{\pm1.4}$ 
\\

$\bigcirc$ & $\checkmark$
& $70.3_{\pm1.0}$ & $51.6_{\pm1.2}$ 
& $59.5_{\pm2.1}$ & $74.9_{\pm1.6}$ 
\\

$\checkmark$ & $\times$
& $68.3_{\pm0.9}$ & $48.1_{\pm2.0}$ 
& $58.3_{\pm1.8}$ & $72.7_{\pm1.9}$ 
\\

\rowcolor{MaDeSRow}
$\checkmark$ & $\checkmark$
& \bm{$73.4_{\pm1.1}$} & \bm{$53.7_{\pm2.5}$} 
& \bm{$61.2_{\pm0.9}$} & \bm{$78.3_{\pm1.3}$} 
\\

\bottomrule
\end{tabular}
\end{table}

We isolate the contribution of preservation and discrimination losses on DVS128 Gesture under $\text{SPC} \times \text{EPS} = 10 \times 128$. We also include the setting of replacing $\mathcal{L}_{\mathrm{pres}}$ with standard DM matching (marked as $\bigcirc$) to isolate the contribution of response alignment and distribution matching.
As shown in \cref{tab:ablation}, adding standard DM to $\mathcal{L}_{\mathrm{disc}}$ improves accuracy, and replacing DM with $\mathcal{L}_{\mathrm{pres}}$ improves it further.
Removing either loss reduces accuracy across all evaluated architectures, with the full objective achieving the best performance across the evaluated architectures. The results indicate that preserving local event structure and reinforcing discriminative cues provide complementary supervision.

\begin{figure*}[t]
    \centering
    \includegraphics[width=17cm]{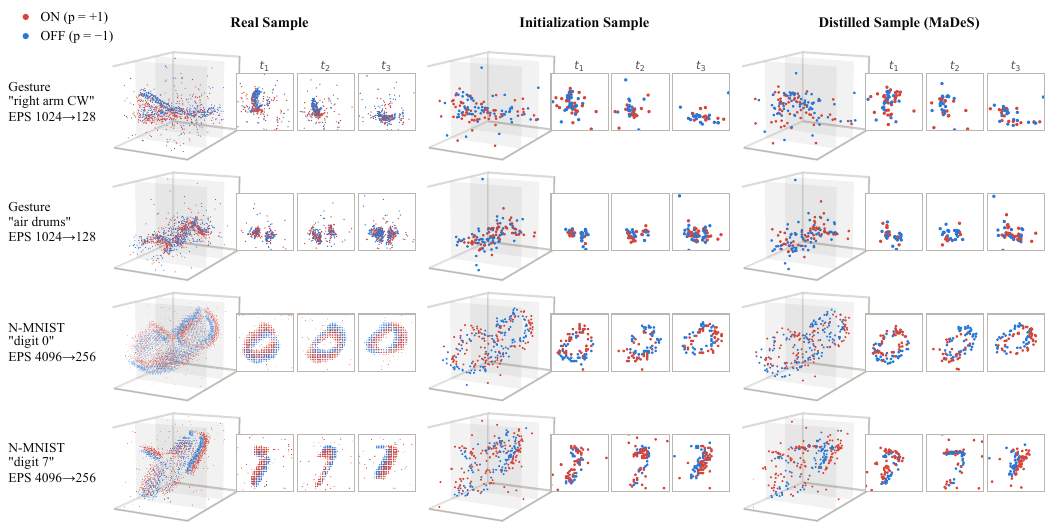}
    \caption{Qualitative comparison between the real sample, distillation initialization, and distilled sample from MaDeS. Each sample is shown as a 3D $(x, t, y)$ event cloud and three equal-duration time slices projected onto the image plane. Red and blue points denote ON and OFF events. For visual clarity, samples with different EPS are drawn with different point sizes.}
    \label{fig:visual}
\end{figure*}

\subsection{Ablation on Model Maturity}
We evaluate the effect of model maturity both for the encoder and the classifier on DVS128 Gesture under a distillation budget of $10 \times 128$. 
We compare models freshly initialized at every distillation step (Fresh), fully trained teachers (Teacher), and classifiers drawn from the maturity-selected state pool $\mathcal{P}$ (Pool).
As shown in \cref{tab:encoder}, with a fixed classifier source,  a freshly initialized matching encoder improves the performance in most comparisons. 
Fully trained teacher classifiers consistently outperform randomly initialized classifiers, and the state pool generally achieves the best performance. 
The combination of a freshly initialized encoder and classifiers of varying maturity achieves the best performance, supporting the proposed maturity-decoupled supervision.
We also provide an ablation study on pool composition in the supplementary material.

\begin{table}[t]
\centering
\footnotesize
\setlength{\tabcolsep}{3pt}
\caption{Effect of maturity for distribution preservation and discrimination on DVS128 Gesture under $\text{SPC} \times \text{EPS} = 10 \times 128$. ``Fresh'' denotes freshly initialized models,  ``Teacher'' denotes fully trained models, and ``Pool'' denotes models from the state pool. The datasets are distilled with SECNet. The best values are highlighted in bold.}
\label{tab:encoder}
\begin{tabular}{ll|cccc}
\toprule
Encoder & Classifier & SECNet & EM & Voxel & VGG-SNN 
\\
\midrule

Teacher & Fresh 
& $64.1_{\pm1.1}$ & $44.4_{\pm1.5}$ 
& $58.6_{\pm2.5}$ & $69.9_{\pm1.4}$ 
\\

Teacher & Teacher 
& $64.7_{\pm1.6}$ & $45.0_{\pm1.8}$ 
& $59.8_{\pm3.9}$ & $71.3_{\pm1.8}$ 
\\

Teacher & Pool 
& $68.9_{\pm0.7}$ & $47.2_{\pm1.9}$ 
& $59.3_{\pm3.3}$ & $72.7_{\pm1.5}$ 
\\

Fresh & Fresh 
& $67.5_{\pm1.1}$ & $47.8_{\pm2.2}$ 
& $57.0_{\pm2.5}$ & $72.5_{\pm1.9}$ 
\\

Fresh & Teacher 
& $69.6_{\pm1.3}$ & $48.0_{\pm2.4}$ 
& $60.3_{\pm1.4}$ & $74.4_{\pm2.3}$ 
\\

\rowcolor{MaDeSRow}
Fresh & Pool 
& \bm{$73.4_{\pm1.1}$} & \bm{$53.7_{\pm2.5}$} 
& \bm{$61.2_{\pm0.9}$} & \bm{$78.3_{\pm1.3}$} 
\\
\bottomrule

\end{tabular}
\end{table}


\subsection{Sensitivity to the Discriminative Weight  \texorpdfstring{$\lambda$}{lambda}}
We evaluate the sensitivity of MaDeS to $\lambda$ by varying the weight of the discriminative term $\lambda$ across different datasets under $\text{SPC} \times \text{EPS} = 10 \times 128$.
As shown in \cref{fig:lambda}, accuracy is relatively stable over a range of intermediate weights and tends to decline at the extremes. 
The preferred range varies across datasets, possibly reflecting differences in the number of classes and the difficulty of distinguishing between classes. 
Datasets with more confusable classes require stronger discrimination guidance to retain useful class cues after compression. We therefore use dataset-specific values of $\lambda$ in the main experiments as listed in \cref{sec:setup}. 
We also provide the evaluation under different distillation budgets in the supplementary material.

\subsection{Computational Cost}
\label{sec:cost}

\begin{table}[t]
\centering
\footnotesize
\setlength{\tabcolsep}{3pt}
\caption{Computational cost on a single H100 GPU across distillation budgets. Preprocessing includes the training of the state pool for MaDeS.}
\label{tab:cost}
\begin{tabular}{l|cc|cc|cc}
\toprule
Method 
& \multicolumn{2}{c|}{Preprocess (min)} 
& \multicolumn{2}{c|}{Distill (min)} 
& \multicolumn{2}{c}{Peak Memory (GB) } 
\\

SPC$\times$EPS 
& 1×128 & 10×256
& 1×128 & 10×256
& 1×128 & 10×256
\\
\midrule
DM 
& \multicolumn{2}{c|}{--}
& 1.5 & 2.6
& 2.05 & 3.62
\\

IDM 
& \multicolumn{2}{c|}{--} 
& 21.7 & 38.3
& 3.16 & 5.40
\\

SADM 
& \multicolumn{2}{c|}{--}
& 3.4 & 4.5
& 2.10 & 3.74
\\

\rowcolor{MaDeSRow}
MaDeS 
& 6.0 & 9.0
& 4.3 & 5.7
& 2.11 & 3.75
\\

\bottomrule
\end{tabular}
\end{table}

We compare the computational cost of MaDeS and three distillation baselines on DVS128 Gesture across distillation budgets. 
We report preprocessing time, distillation time, and peak memory, averaged over three runs on a single H100 GPU. 
As shown in \cref{tab:cost}, MaDeS incurs a one-time cost to construct the state pool, which can be reused across SPC budgets for the same dataset and EPS.
IDM achieves competitive accuracy, but training its model pool online substantially increases distillation time, which grows with EPS.
Overall, MaDeS achieves improved performance with a one-time preprocessing cost and modest additional distillation time.

\subsection{Visualization and Analysis}
\label{sec:vis}
We visualize real samples, their randomly subsampled initializations, and the corresponding distilled samples for two classes each from DVS128 Gesture and N-MNIST. 
\cref{fig:visual} shows the 3D event clouds and their time slices projected onto the image plane. 
The initializations retain the overall shape of the real samples despite using fewer events. 
The distilled samples have modified individual events while largely preserving the real spatiotemporal layout.
In short, MaDeS represents each sample with far fewer events, keeps it realistic in event space, and makes it substantially more informative for recognition.

\section{Conclusion}
\label{sec:conclusion}
In this paper, we introduced MaDeS for two-axis event dataset distillation, reducing both the number of samples per class and the number of events per sample while retaining the raw event format. 
MaDeS uses freshly initialized encoders to preserve the original event distribution and trained classifiers drawn from a maturity-selected pool to retain class-discriminative cues, serving as maturity-decoupled supervision.
Extensive experiments across various settings validate the effectiveness and cross-architecture generalizability of the distilled datasets and show the complementary roles of the two objectives.  

\par

Nevertheless, MaDeS uses dataset-specific weights for the discrimination loss. 
We will consider self-adaptive weighting strategies based on distillation gradients. MaDeS also assigns the same number of events to every synthetic sample. 
Allocating different numbers of events per sample for simple and complex motions may improve distillation under a fixed budget. 
Finally, MaDeS builds state pools for each EPS, while training classifiers on samples with varying event counts may produce state pools that can serve multiple EPS settings.

{\small
\bibliographystyle{ieeenat_fullname}
\bibliography{main_clean}

@article{wang2018dd,
  title={Dataset Distillation},
  author={Wang, Tongzhou and Zhu, Jun-Yan and Torralba, Antonio and Efros, Alexei A.},
  journal={arXiv preprint arXiv:1811.10959}, year={2018}
}

@article{gao2023thu1,
  title={Action Recognition and Benchmark Using Event Cameras},
  author={Gao, Yue and Lu, Jiaxuan and Li, Siqi and Ma, Nan and Du, Shaoyi and Li, Yipeng and Dai, Qionghai},
  journal={IEEE Transactions on Pattern Analysis and Machine Intelligence},
  year={2023},
  volume={45},
  number={12},
  pages={14081-14097},
}

@article{gao2024thu2,
  title={Hypergraph-Based Multi-View Action Recognition Using Event Cameras},
  author={Gao, Yue and Lu, Jiaxuan and Li, Siqi and Li, Yipeng and Du, Shaoyi},
  journal={IEEE Transactions on Pattern Analysis and Machine Intelligence},
  volume={46},
  number={10},
  pages={6610-6622},
  year={2024},
}

@techreport{krizhevsky2009cifar,
  author = {Krizhevsky, Alex and Hinton, Geoffrey},
  title = {Learning multiple layers of features from tiny images},
  institution = {University of Toronto},
  year = {2009}
}

@inproceedings{amir2017gesture,
  title={A Low Power, Fully Event-Based Gesture Recognition System},
  author={Amir, Arnon and Taba, Brian and Berg, David and others},
  booktitle={Proceedings of the IEEE/CVF Conference on Computer Vision and Pattern Recognition},
  year={2017}
}

@article{orchard2015nmnist,
  title={Converting Static Image Datasets to Spiking Neuromorphic Datasets Using Saccades},
  author={Orchard, Garrick and Jayawant, Ajinkya and Cohen, Gregory K. and Thakor, Nitish},
  journal={Frontiers in Neuroscience}, volume={9},
  pages={437},
  year={2015}
}

@inproceedings{ren2025secnet,
  title={Scalable Event Cloud Network for Event-based Classification},
  author={Ren, Hongwei and Ma, Fei and Lin, Xiaopeng and Fang, Yuetong and Huang, Hongxiang and Zhou, Yue and Huang, Yulong and Fu, Haotian and Yang, Ziyi and Jiang, Youxin and Wu, Xiangqian and Cheng, Bojun},
  booktitle={Proceedings of the International Conference on Machine Learning},
  year={2026}
}

@article{ren2024eventmamba,
  title={Rethinking Efficient and Effective Point-Based Networks for Event Camera Classification and Regression},
  author={Ren, Hongwei and Zhou, Yue and Zhu, Jiadong and Lin, Xiaopeng and Fu, Haotian and Huang, Yulong and Fang, Yuetong and Ma, Fei and Yu, Hao and Cheng, Bojun},
  journal={IEEE Transactions on Pattern Analysis \& Machine Intelligence },
  year={2025},
  volume={47},
  pages={6228-6241},
}

@article{sengupta2019vgg,
  author={Sengupta, Abhronil and Ye, Yuting  and Wang, Robert and Liu, Chiao and Roy, Kaushik},
  title={Going Deeper in Spiking Neural Networks: {VGG} and Residual Architectures},
  journal={Frontiers in Neuroscience},
  volume={13},
  year={2019},
}

@inproceedings{welling2009herding,
  title = {Herding dynamical weights to learn},
  author = {Welling, Max},
  booktitle={Proceedings of the 26th International Conference on Machine Learning},
  year = {2009}
}

@inproceedings{sener2018kcenter,
  title = {Active learning for convolutional neural networks: A core-set approach},
  author = {Sener, Ozan and Savarese, Silvio},
  booktitle={Proceedings of the International Conference on Learning Representations},
  year = {2018}
}

@inproceedings{araghi2025density,
  title={Making Every Event Count: Balancing Data Efficiency and Accuracy in Event Camera Subsampling},
  author={Araghi, Hesam and van Gemert, Jan and Tomen, Nergis},
  booktitle={Proceedings of the IEEE/CVF Conference on Computer Vision and Pattern Recognition Workshops},
  year={2025}
}

@inproceedings{girbau2025poled,
  title={Probabilistic Online Event Downsampling},
  author={Girbau-Xalabarder, Andreu and Nagata, Jun and Sumiyoshi, Shinichi}, booktitle={Proceedings of the IEEE/CVF Conference on Computer Vision and Pattern Recognition Workshops},
  year={2025}
}

@inproceedings{zhao2023dm,
  title={Dataset Condensation with Distribution Matching},
  author={Zhao, Bo and Bilen, Hakan},
  booktitle={Proceedings of the IEEE/CVF Winter Conference on Applications of Computer Vision},
  year={2023}
}

@inproceedings{zhao2023idm,
  title={Improved Distribution Matching for Dataset Condensation},
  author={Zhao, Ganlong and Li, Guanbin and Qin, Yipeng and Yu, Yizhou},
  booktitle={Proceedings of the IEEE/CVF Conference on Computer Vision and Pattern Recognition},
  year={2023}
}

@inproceedings{yim2025sadm,
  title={Dataset Distillation of 3D Point Clouds via Distribution Matching},
  author={Yim, Jae-Young and Kim, Dongwook and Sim, Jae-Young},
  booktitle={Proceedings of the Conference on Neural Information Processing Systems},
  year={2025}
}

@inproceedings{liu2025wmdd,
  title={Dataset Distillation via the Wasserstein Metric},
  author={Liu, Haoyang and Li, Yijiang and Xing, Tiancheng and Wang, Peiran and Dalal, Vibhu and Li, Luwei and He, Jingrui and Wang, Haohan},
  booktitle={Proceedings of the IEEE/CVF International Conference on Computer Vision},
  year={2025}
}

@ARTICLE{ghosh2025survey,
  author={Ghosh, Suman and Gallego, Guillermo},
  title={Event-Based Stereo Depth Estimation: A Survey},
  journal={IEEE Transactions on Pattern Analysis and Machine Intelligence},
  year={2025},
  volume={47},
  number={10},
  pages={9130-9149},
}

@ARTICLE{gallego2020survey,
  author={Gallego, Guillermo and Delbrück, Tobi and Orchard, Garrick and Bartolozzi, Chiara and Taba, Brian and Censi, Andrea and Leutenegger, Stefan and Davison, Andrew J. and Conradt, Jörg and Daniilidis, Kostas and Scaramuzza, Davide},
  title={Event-Based Vision: A Survey},
  journal={IEEE Transactions on Pattern Analysis and Machine Intelligence},
  year={2022},
  volume={44},
  number={1},
  pages={154-180},
}

@article{Gehrig2024appli,
  author={Gehrig, Daniel and Davide Scaramuzza},
  title={Low-latency automotive vision with event cameras},
  journal={Nature},
  volume={629},
  number={8014},
  year={2024},
  pages={1034-1040}
}

@inproceedings{Aitsam2024appli,
  author={Aitsam, Muhammad and Davies, Sergio and Di Nuovo, Alessandro},
  title={Event Camera-Based Real-Time Gesture Recognition for Improved Robotic Guidance},
  booktitle={Proceedings of the International Joint Conference on Neural Networks},
  year={2024},
}

@inproceedings{fang2021spike,
 author = {Fang, Wei and Yu, Zhaofei and Chen, Yanqi and Huang, Tiejun and Masquelier, Timoth\'{e}e and Tian, Yonghong},
 booktitle = {Proceedings of the Conference on Neural Information Processing Systems},
 title = {Deep Residual Learning in Spiking Neural Networks},
 year={2021}
}

@inproceedings{perot2020learning,
  title={Learning to Detect Objects with a 1 Megapixel Event Camera},
  author={Perot, Etienne and de Tournemire, Pierre and Nitti, Davide and Masci, Jonathan and Sironi, Amos},
  booktitle={Proceedings of the Conference on Neural Information Processing Systems},
  year={2020}
}

@inproceedings{gu2021eventdrop,
  title={{EventDrop}: Data Augmentation for Event-based Learning},
  author={Gu, Fuqiang and Sng, Weicong and Hu, Xuke and Yu, Fangwen},
  booktitle={Proceedings of the International Joint Conference on Artificial Intelligence},
  year={2021}
}

@article{ye2025pace,
  title={Learning from Dense Events: Towards Fast Spiking Neural Networks Training via Event Dataset Distillation},
  author={Ye, Shuhan and Yu, Yi and Zhang, Qixin and Kong, Chenqi and Wu, Qiangqiang and Wang, Kun and Jiang, Xudong},
  journal={arXiv preprint arXiv:2511.12095},
  year={2025}}

@inproceedings{zhang2024dance,
  title={{DANCE}: Dual-View Distribution Alignment for Dataset Condensation},
  author={Zhang, Hansong and Li, Shikun and Lin, Fanzhao and Wang, Weiping and Qian, Zhenxing and Ge, Shiming},
  booktitle={Proceedings of the International Joint Conference on Artificial Intelligence},
  year={2024}
}

@article{lei2023ddsurvey,
  title={A Comprehensive Survey of Dataset Distillation},
  author={Lei, Shiye and Tao, Dacheng},
  journal={IEEE Transactions on Pattern Analysis and Machine Intelligence},
  volume={46},
  number={1},
  pages={17--32},
  year={2023},
}

@inproceedings{yang2025objClassi,
    author={Yang, Yan and Pan, Liyuan and Li, Dongxu and Liu, Liu},
    title={{EZSR}: Event-based Zero-Shot Recognition},
    booktitle={Proceedings of the IEEE/CVF Conference on Computer Vision and Pattern Recognition},
    year={2025},
}

@InProceedings{cao2026humanAct,
    author={Cao, Meiqi and Zhang, Jiachao and Jiang, Xin and Yan, Rui and Yao, Yazhou and Li, Zechao and Shu, Xiangbo},
    title={Seeing Motion Through Polarity for Event-based Action Recognition},
    booktitle={Proceedings of the IEEE/CVF Conference on Computer Vision and Pattern Recognition},
    year={2026},
}

@article{zhang2026fineMotion,
  author = {Yi Zhang and Xiuping Liu and Jun Zhou and Yang Liu and Hongchen Tan},
  title = {Semantics-aware high-frequency enhancement for event-based lip-reading},
  journal = {Information Sciences},
  volume = {742},
  pages = {123026},
  year = {2026}
}

@inproceedings{he2016resnet,
 title={Deep Residual Learning for Image Recognition},
 author={He, Kaiming and Zhang, Xiangyu and Ren, Shaoqing and Sun, Jian},
 booktitle={Proceedings of the IEEE Conference on Computer Vision and Pattern Recognition},
 year={2016}
}

@inproceedings{gehrig2019est,
  title={End-to-End Learning of Representations for Asynchronous Event-Based Data},
  author={Gehrig, Daniel and Loquercio, Antonio and Derpanis, Konstantinos G. and Scaramuzza, Davide},
  booktitle= {Proceedings of the IEEE/CVF International Conference on Computer Vision},
  year={2019}
}

@inproceedings{sironi2018hats,
  title={{HATS}: Histograms of Averaged Time Surfaces for Robust Event-Based Object Classification},
  author={Sironi, Amos and Brambilla, Manuele and Bourdis, Nicolas and Lagorce, Xavier and Benosman, Ryad},
  booktitle={Proceedings of the IEEE/CVF Conference on Computer Vision and Pattern Recognition},
  year={2018}
}

@inproceedings{qi2017pointnet,
  title={PointNet: Deep Learning on Point Sets for 3D Classification and Segmentation},
  author={Qi, Charles R. and Su, Hao and Mo, Kaichun and Guibas, Leonidas J.},
  booktitle={Proceedings of the IEEE/CVF Conference on Computer Vision and Pattern Recognition},
  year={2017}
  }

@inproceedings{qi2017pointnet++,
  title={PointNet++: Deep Hierarchical Feature Learning on Point Sets in a Metric Space},
  author={Qi, Charles R. and Yi, Li and Su, Hao and Guibas, Leonidas J.},
  booktitle={Proceedings of the Conference on Neural Information Processing Systems},
  year={2017}
}

@inproceedings{zhao2021dc,
  title={Dataset Condensation with Gradient Matching},
  author={Zhao, Bo and Mopuri, Konda Reddy and Bilen, Hakan},
  booktitle={Proceedings of the International Conference on Learning Representations},
  year={2021}
}

@inproceedings{zhao2021dsa,
  title={Dataset condensation with Differentiable Siamese Augmentation},
  author={Zhao, Bo and Bilen, Hakan},
  booktitle={Proceedings of the International Conference on Machine Learning},
  year={2021}
}

@inproceedings{cazenavette2022mtt,
  title={Dataset Distillation by Matching Training Trajectories},
  author={Cazenavette, George and Wang, Tongzhou and Torralba, Antonio and Efros, Alexei A. and Zhu, Jun-Yan},
  booktitle={Proceedings of the IEEE/CVF Conference on Computer Vision and Pattern Recognition},
  year={2022}
}

@inproceedings{wang2022cafe,
  title={{CAFE}: Learning to Condense Dataset by Aligning Features},
  author={Wang, Kai and Zhao, Bo and Peng, Xiangyu and Zhu, Zheng and Yang, Shuo and Wang, Shuo and Huang, Guan and Bilen, Hakan and Wang, Xinchao and You, Yang},
  booktitle={Proceedings of the IEEE/CVF Conference on Computer Vision and Pattern Recognition},
  year={2022}
}

@inproceedings{zhang2024m3d,
  title={{M3D}: Dataset Condensation by Minimizing Maximum Mean Discrepancy},
  author={Zhang, Hansong and Li, Shikun and Wang, Pengju and Zeng, Dan and Ge, Shiming},
  booktitle={Proceedings of the AAAI Conference on Artificial Intelligence},
  year={2024}
}

@inproceedings{wang2025ncfm,
  title={Dataset Distillation with Neural Characteristic Function: A Minmax Perspective},
  author={Wang, Shaobo and Yang, Yicun and Liu, Zhiyuan and Sun, Chenghao and Hu, Xuming and He, Conghui and Zhang, Linfeng},
  booktitle={Proceedings of the IEEE/CVF Conference on Computer Vision and Pattern Recognition},
  year={2025}
}

@inproceedings{yin2023sre2l,
  title={Squeeze, Recover and Relabel: Dataset Condensation at {ImageNet} Scale From A New Perspective},
  author={Yin, Zeyuan and Xing, Eric and Shen, Zhiqiang},
  booktitle={Proceedings of the Conference on Neural Information Processing Systems},
  year={2023}
}

@inproceedings{bo2025point,
  title={Point Cloud Dataset Distillation},
  author={Bo, Deyu and Wang, Xinchao},
  booktitle={Proceedings of the International Conference on Machine Learning},
  year={2025}
}

@inproceedings {carlini2027repara,
  author={Carlini, Nicholas and Wagner, David},
  booktitle={Proceedings of the IEEE Symposium on Security and Privacy},
  title={Towards Evaluating the Robustness of Neural Networks},
  year={2017},
}

@inproceedings{guo2024datm,
  title={Towards Lossless Dataset Distillation via Difficulty-Aligned Trajectory Matching},
  author={Guo, Ziyao and Wang, Kai and Cazenavette, George and Li, Hui and Zhang, Kaipeng and You, Yang},
  booktitle={Proceedings of the International Conference on Learning Representations},
  year={2024}
}

@INPROCEEDINGS{zhu2019voxel,
  title={Live Demonstration: Unsupervised Event-Based Learning of Optical Flow, Depth and Egomotion}, 
  author={Zhu, Alex Zihao and Yuan, Liangzhe and Chaney, Kenneth and Daniilidis, Kostas},
  booktitle={Proceedings of the IEEE/CVF Conference on Computer Vision and Pattern Recognition Workshops}, 
  year={2019},
}

@article{li2024pruning,
  title = {Dataset Distillation Using Parameter Pruning},
  author = {Li, Guang and Togo, Ren and Ogawa, Takahiro and Haseyama, Miki},
  journal = {IEICE Transactions on Fundamentals of Electronics, Communications and Computer Sciences},
  volume = {E107-A},
  number = {6},
  pages = {936--940},
  doi = {10.1587/transfun.2023EAL2053},
  year = {2024}
}

@article{li2024iadd,
  title = {Importance-Aware Adaptive Dataset Distillation},
  author = {Li, Guang and Togo, Ren and Ogawa, Takahiro and Haseyama, Miki},
  journal = {Neural Networks},
  pages = {106154},
  doi = {10.1016/j.neunet.2024.106154},
  year = {2024}
}

@inproceedings{li2025hyperbolic,
  title = {Hyperbolic Dataset Distillation},
  author = {Li, Wenyuan and Li, Guang and Maeda, Keisuke and Ogawa, Takahiro and Haseyama, Miki},
  booktitle = {Advances in Neural Information Processing Systems},
  url = {https://guang000.github.io/HDD-Webpage/},
  year = {2025}
}

@article{ma2026difficulty,
  title = {Difficulty-Aware Statistical Matching for Dataset Distillation},
  author = {Ma, Hongxu and Li, Guang and Wang, Shijie and Zhou, Dongzhan and Yang, Suorong and Sun, Baoli and Ogawa, Takahiro and Haseyama, Miki and Wang, Zhihui},
  journal = {arXiv preprint arXiv:2609.34299},
  url = {https://arxiv.org/abs/2609.34299},
  year = {2026}
}

@inproceedings{peng2026clpdd,
  title     = {Closed-Form Linear-Probe Dataset Distillation for Pre-trained Vision Models},
  author    = {Peng, Bincheng and Li, Guang and Liu, Ping and Ogawa, Takahiro and Haseyama, Miki},
  booktitle = {Advances in Neural Information Processing Systems},
  year      = {2026}
}

@article{li2026srg,
  title = {Self-Supervised Representation-Guided Generative Dataset Distillation},
  author = {Li, Mingzhuo and Li, Guang and Ye, Linfeng and Mao, Jiafeng and Ogawa, Takahiro and Plataniotis, Konstantinos N. and Haseyama, Miki},
  journal = {arXiv preprint arXiv:2608.03218},
  year = {2026},
  url = {https://arxiv.org/abs/2608.03218}
}

@article{li2022soft,
  title = {Compressed Gastric Image Generation Based on Soft-Label Dataset Distillation for Medical Data Sharing},
  author = {Li, Guang and Togo, Ren and Ogawa, Takahiro and Haseyama, Miki},
  journal = {Computer Methods and Programs in Biomedicine},
  volume = {227},
  pages = {107189},
  doi = {10.1016/j.cmpb.2022.107189},
  year = {2022}
}

@inproceedings{ma2026fd2,
  title = {{FD$^2$}: A Dedicated Framework for Fine-Grained Dataset Distillation},
  author = {Ma, Hongxu and Li, Guang and Wang, Shijie and Zhou, Dongzhan and Sun, Baoli and Ogawa, Takahiro and Haseyama, Miki and Wang, Zhihui},
  booktitle = {European Conference on Computer Vision},
  url = {https://arxiv.org/abs/2603.25144},
  year = {2026}
}

@inproceedings{li2024global,
  title = {Generative Dataset Distillation: Balancing Global Structure and Local Details},
  author = {Li, Longzhen and Li, Guang and Togo, Ren and Maeda, Keisuke and Ogawa, Takahiro and Haseyama, Miki},
  booktitle = {Proceedings of the IEEE/CVF Conference on Computer Vision and Pattern Recognition Workshops},
  pages = {7664--7671},
  doi = {10.1109/CVPRW63382.2024.00762},
  year = {2024}
}

@article{li2026foreground,
  title = {Foreground-Aware Dataset Distillation via Dual-Path Patch Selection},
  author = {Li, Mingzhuo and Li, Guang and Li, Longzhen and Togo, Ren and Maeda, Keisuke and Ogawa, Takahiro and Haseyama, Miki},
  journal = {arXiv preprint arXiv:2601.02727},
  year = {2026},
  url = {https://arxiv.org/abs/2601.02727}
}

@inproceedings{fan2026deco,
  title = {{DeCO}: Discriminative Evidence Composition for Fine-Grained Dataset Distillation},
  author = {Fan, Chuixuan and Li, Guang and Wang, Shijie and Zhou, Dongzhan and Sun, Baoli and Ogawa, Takahiro and Haseyama, Miki and Wang, Zhihui},
  booktitle = {NeurIPS Workshop on Foundations of Efficient Deep Learning},
  url = {https://arxiv.org/abs/2608.25480},
  year = {2026}
}

@inproceedings{li2025sampling,
  title = {Task-Specific Generative Dataset Distillation with Difficulty-Guided Sampling},
  author = {Li, Mingzhuo and Li, Guang and Mao, Jiafeng and Ye, Linfeng and Ogawa, Takahiro and Haseyama, Miki},
  booktitle = {Proceedings of the IEEE/CVF International Conference on Computer Vision Workshops},
  pages = {6853--6860},
  doi = {10.1109/ICCVW69036.2025.00708},
  year = {2025}
}

@article{li2026sas,
  title = {{SAS}: Semantic-Aware Sampling for Generative Dataset Distillation},
  author = {Li, Mingzhuo and Li, Guang and Ye, Linfeng and Mao, Jiafeng and Ogawa, Takahiro and Plataniotis, Konstantinos N. and Haseyama, Miki},
  journal = {IEEE Open Journal of Signal Processing},
  year = {2026},
  url = {https://arxiv.org/abs/2605.18012},
  volume = {7},
  pages = {668--677}
}

@article{ren2026protoblend,
  title = {Efficient Video Dataset Distillation via Cluster-Guided Prototype Blending},
  author = {Ren, Chongle and Li, Guang and Huang, Wenbo and Saito, Naoki and Ogawa, Takahiro and Haseyama, Miki},
  journal = {arXiv preprint arXiv:2608.03269},
  year = {2026},
  url = {https://arxiv.org/abs/2608.03269}
}

@article{tian2026motion,
  title = {Latent Dataset Distillation for Human Motion Prediction},
  author = {Tian, Ge and Li, Guang and Ogawa, Takahiro and Haseyama, Miki},
  journal = {arXiv preprint arXiv:2609.26430},
  year = {2026},
  url = {https://arxiv.org/abs/2609.26430}
}

@misc{li2022awesome,
  author={Li, Guang and Zhao, Bo and Wang, Tongzhou},
  title={Awesome Dataset Distillation},
  howpublished={\url{https://github.com/Guang000/Awesome-Dataset-Distillation}},
  year={2022}
}

@inproceedings{li2020soft,
  title={Soft-Label Anonymous Gastric X-Ray Image Distillation},
  author={Li, Guang and Togo, Ren and Ogawa, Takahiro and Haseyama, Miki},
  booktitle={Proceedings of the IEEE International Conference on Image Processing},
  pages={305--309},
  year={2020}
}
}

\clearpage
\appendix
\section*{Appendix}
\section{Detailed Experimental Settings}
\label{sec:supp_settings}
We expand on the experimental setup described in the main paper by providing further details on the datasets and the implementation. For simplicity, we omit repeated citations to datasets and methods in the supplementary tables. 

\paragraph{Datasets.}
We evaluate on four event recognition datasets. The preprocessing from recordings to samples follows the settings of the SECNet~\cite{ren2025secnet}.
DVS128 Gesture~\cite{amir2017gesture} is recorded with a DVS128 sensor ($128\times128$) and contains 11 hand and arm gestures (\eg, hand clapping, arm rotations, air guitar) performed by 29 subjects under three lighting conditions.
The ``random other gestures'' class is removed, leaving 10 classes and 1{,}342 labelled gesture segments, split by subject into 1{,}078 training and 264 test segments. For recognition, each segment is cut into 0.5\,s windows with a 0.25\,s stride. Each window is treated as one sample, leaving 26816 train samples and 6959 test samples.

\par

THU-EACT-50~\cite{gao2023thu1,gao2024thu2} is recorded with a CeleX-V sensor ($1280\times800$) and contains 50 daily human actions performed by 105 subjects with 8{,}500/2{,}000 train/test recordings. Each recording covers one whole action of about 2--5\,s with $\sim$1--2\,M events, and is treated as one sample.

\par

N-MNIST~\cite{orchard2015nmnist} is recorded by moving an ATIS sensor ($34\times34$ region) in three saccades in front of a monitor showing MNIST digits (10 classes), so the events are produced by the sensor's own motion over a static image.
One recording ($\approx$ 300\,ms) is treated as one sample, leaving 60{,}000 train samples and 10{,}000 test samples.

\par

N-Caltech101~\cite{orchard2015nmnist} is recorded in the same way from the Caltech101 images (101 classes, 8{,}709 recordings, $240\times180$ region).
Three fixed 30\,ms windows are taken per recording, one in each saccade (35--65, 135--165 and 235--265\,ms), and duplicate events at 1\,ms resolution are removed.
The split is at the recording level (80/20 per class), giving 6{,}967 training and 1{,}742 test recordings, and 20{,}901 train and 5{,}226 test samples.

\par

\begin{table}[t]
\centering
\footnotesize
\caption{Dataset splits and standard events per sample (EPS).}
\label{tab:sup_dataset}
\begin{tabular}{lcccc}
\toprule
Dataset & Classes & Train & Test & Standard EPS \\
\midrule
DVS128 Gesture 
& 10 & 26{,}816 & 6{,}959 & 1{,}024 
\\

N-MNIST 
& 10 & 60{,}000 & 10{,}000 & 4{,}096 
\\

N-Caltech101 
& 101 & 20{,}901 & 5{,}226 & 10{,}240 
\\

THU-EACT-50
& 50 & 8{,}500 & 2{,}000 & 4{,}096 
\\

\bottomrule
\end{tabular}
\end{table}

\paragraph{Event Recognition Networks.}
We include four event recognition networks from three families.
Every network receives the same input type (\eg the sample) and converts it into its own representation.
SECNet~\cite{ren2025secnet} is a point-based network that treats each event as a 4D point. It builds a three-stage hierarchy in which farthest point sampling (FPS) reduces the events to $\text{EPS}/2$, $\text{EPS}/4$ and $\text{EPS}/8$ group centres, aggregates each group from its nearest neighbours in feature space, and models spatial and temporal structure with separate aggregation layers. As the original configuration targets 1024 or more events, we scale the number of groups with EPS.

\par

EventMamba~\cite{ren2024eventmamba} is also point-based, with the same FPS-based downsampling, but differs from SECNet in most other components: it ignores polarity, groups neighbours in coordinate space, aggregates the time-ordered groups with a bidirectional Mamba scan, and pools with attention.

\par

ResNet-18 ~\cite{he2016resnet} for voxel grids~\cite{zhu2019voxel} represents the dense methods. The events are converted into a voxel grid with 8 time bins over the time span of each sample, in which each event adds its polarity to the two nearest time bins with linear weights, at a spatial resolution of $32 \times 32$. The time bins are used as input channels of a standard 2D ResNet-18, whose stem is replaced by a $3 \times 3$ stride-1 convolution without max-pooling to suit the small input.

\par

VGG-SNN~\cite{sengupta2019vgg} represents spike methods. The events are accumulated into $T = 8$ frames of $48 \times 48$ with separate ON and OFF channels (event counts per pixel), which are fed to the network over 8 time steps. The prediction is averaged over time.

\par

Point-based networks operate directly on individual events and are therefore well suited to raw event dataset distillation. We use SECNet as the default distillation network and use the other networks primarily for cross-architecture evaluation. 

\par

The voxel-based and spiking networks use all events in each sample and discretize their spatial coordinates and timestamps. Moving an event within the same pixel and time bin therefore leaves the input representation unchanged.
The point-based networks inherently downsample the events within each sample. We follow SECNet~\cite{ren2025secnet} in setting the target number of events per sample (standard EPS) except that THU-EACT-50 is not included and we set the standard EPS to 4{,}096 based on preliminary experiments. We summarize the dataset splits and standard EPS values in \Cref{tab:sup_dataset}.

\begin{table}[t]
\centering
\footnotesize
\caption{Default settings for distillation, state pool construction, and evaluation.}
\label{tab:sup_dis_set}
\setlength{\tabcolsep}{2.8pt}

\begin{tabular}{ll}
\toprule
\multicolumn{2}{l}{\emph{Distillation}} 
\\

Distillation optimizer & Adam \\
Optimizer Settings & lr $=3\times10^{-3}$
\\

Iterations & 2{,}500 \\

Real batch $B$ & 64 (8 on THU-EACT-50 and N-Caltech101) 
\\

Distillation network & SECNet 
\\
Top-1 response weight $\gamma$ & $0.5$ \\
Discrimination weight $\lambda$ & $0.01$ on Gesture and THU-EACT-50 \\
& $0.003$ on N-MNIST; $0.1$ on N-Caltech101 \\

\midrule

\multicolumn{2}{l}{\emph{State pool}} 
\\

Training optimizer & AdamW \\
Optimizer settings & lr $=10^{-3}$, batch $=128$ \\
 & (batch =$64$ on N-MNIST) \\
 
Training iterations & 6k on Gesture, N-MNIST; \\ 
 & 15k on THU-EACT-50; \\
 & 60k on N-Caltech101 \\
Number of trajectories $R$ & 3, each with a different seed 
\\
Number of checkpoints $K$ & 20 
\\
Target fractions $q$ & $q\in\{0.2,0.4,0.6,0.8\}$, \\
 & resulting in 12 classifier heads in total \\

\midrule

\multicolumn{2}{l}{\emph{Evaluation}} 
\\
Evaluation optimizer & AdamW \\
Optimizer settings & lr $=10^{-3}$, wd $=10^{-4}$, cosine \\

Iterations & 3{,}000 \\

Batch size & 64 \\

Loss & cross-entropy; label smoothing $=0.2$ \\

Repetitions & 3 distilled sets $\times$ 3 training seeds \\

\bottomrule
\end{tabular}
\end{table}
\begin{algorithm}[t]
\caption{Two-Axis Event Dataset Distillation with Maturity-Decoupled Supervision (MaDeS)}
\label{alg}
{\footnotesize \itshape Equation numbers refer to the main paper.}

\begin{algorithmic}[1]
\small
\REQUIRE $\mathcal D$: original event dataset;
SPC: number of samples per class in the distilled dataset;
EPS: number of events per sample in the distilled dataset;
$\mathcal{S}(E,n)$: subsample operator;
$\mathbf{u}=(u_x, u_y, u_t, u_p)$: learnable unconstrained vector;
$R$: training runs when constructing the state pool; 
$K$: checkpoints per run when constructing the state pool;
$\mathcal{Q}$: target maturity fractions;
$B$: real batch size per class;
$\gamma$: weight to balance the top-1 response;
$\lambda$: weight to balance the discrimination term.
\ENSURE $\tilde{\mathcal{D}}$: distilled event dataset.

\STATE \textbf{Stage 1: State pool construction}
\STATE Normalize $(x,y,t)$ in each real window to $[0,1]$
\STATE Initialize $\mathcal P\leftarrow\varnothing$
\FOR{$r=1,\ldots,R$}
    \STATE Obtain the subsampled dataset $\mathcal{D}^*$ by conducting $\mathcal{S}(E, \text{EPS})$ for each sample $E$ in $\mathcal{D}$
    \STATE Train a classifier on $\mathcal{D}^*$ and save $K$ checkpoints. 
    \STATE Calculate the maturity $s_{r,j}$ of each checkpoint $j$
    \FOR{each $q \in \mathcal{Q}$}
        \STATE Select $j_{r,q}$ by Eq.~(6)
        \STATE Add the corresponding checkpoint to $\mathcal{P} $
    \ENDFOR
\ENDFOR

\STATE \textbf{Stage 2: Raw event distillation}
\FOR{each class $c=1,\ldots,C$}
    \STATE Draw $\text{SPC}$ real samples from $\mathcal{D}$
    \STATE Process the samples with $\mathcal S(E,\mathrm{EPS})$
    \STATE Initialize the learnable vectors $\mathbf{u}$ via Eq.~(1)
\ENDFOR

\FOR{each distillation step}
    \STATE Map $\mathbf{u}$ to synthetic events by Eq.~(1) and sort each $\tilde E_{c,i}$ by timestamp
    \STATE Randomly initialize the encoder $f_{\theta_{\mathrm{init}}}$ 
    \STATE Sample a classifier $\theta_{\mathrm{disc}}$ uniformly from $\mathcal{P}$
    \FOR{each class $c=1,\ldots,C$}
        \STATE Draw $B$ real samples from $\mathcal{D}$ and form $\mathcal B_c^*$ by Eq.~(2). 
        \STATE Apply the same augmentation to real and synthetic samples.
        \FOR{each encoder stage $s=1,\ldots,S$}
            \STATE Use $f^s_{\theta_{\mathrm{init}}}$ to obtain the ranked descriptors $\mathcal Z_c^s,\tilde{\mathcal Z}_c^s$ by Eq.~(3)
            \STATE Extract the top-1 descriptors $\mathcal M_c^s,\tilde{\mathcal M}_c^s$ by Eq.~(4)
        \ENDFOR
    \ENDFOR
    
    \STATE Compute $\mathcal L_{\mathrm{pres}}$ over all classes and stages by Eq.~(5)
    \STATE Compute $\mathcal L_{\mathrm{disc}}$ from the logits of $h_{\theta_{\mathrm{disc}}}$ on the synthetic samples by Eq.~(7)
    \STATE Update $\mathbf u$ using $\nabla_{\mathbf u} \ \mathcal L_{\mathrm{MaDeS}}$ from Eq.~(8)
\ENDFOR

\STATE Form $\tilde{\mathcal D}$ by mapping the updated parameters $\mathbf{u}$ via Eq.~(1) and sorting each window by timestamp

\RETURN $\tilde{\mathcal D}$
\end{algorithmic}

\end{algorithm}

\paragraph{Baseline Implementation.}
We compare MaDeS with sample selection, event selection, and dataset distillation baselines in our experiments.

\par

Sample selection chooses whole samples within each class. Random draws samples uniformly, and Closest selects those nearest to the class mean in feature space. Herding~\cite{welling2009herding} greedily selects samples to approximate the class means. K-Center~\cite{sener2018kcenter} starts with the sample nearest to the mean and repeatedly adds the sample farthest from the selected set. For feature-based selection, we use a SECNet trained on the full training set at the target EPS as the feature extractor.

\par

Event selection subsamples each sample independently and preserves timestamp order. Stride selects events at regular intervals in the ordered stream. Farthest point sampling (FPS)~\cite{qi2017pointnet++} operates in normalized $(x,y,t)$ space.
Araghi \etal~\cite{araghi2025density} assign higher retention probabilities to events in regions of high spatiotemporal event density. POLED~\cite{girbau2025poled} derives each event's retention probability from an estimated spatial probability density function. Both methods use these probabilities to accept or reject individual events, leaving variable numbers of retained events across samples. In our experiments, we use these probabilities as sampling weights to select exactly EPS events from each sample.

\par

When evaluating either selection axis alone, we randomly subsample events or draw samples for the other axis. We also evaluate combinations of the two selections in \cref{sec:selection}.

\par

For distillation baselines, we follow the published settings with adaptations for raw events. We omit IDM's image partition-and-expansion augmentation and disable SADM's learned rotations to preserve the sensor coordinate frame. For WMDD, we omit the image-specific total variation and pixel $\ell_2$ priors to ensure equal comparison with other methods.
For MaDeS, a repulsion term is added to prevent synthetic events from collapsing onto each other. 

\begin{table}[t]
\centering
\footnotesize
\setlength{\tabcolsep}{4pt}
\caption{Cross-architecture accuracy (\%) on DVS128 Gesture across budgets using SECNet for distillation.}
\label{tab:sup_cross_ipc}
\begin{tabular}{l|cccc|c}
\toprule
Method & SECNet & EventMamba & Voxel & SNN & Mean \\
\midrule

\multicolumn{6}{l}{\textbf{$\text{SPC}\times\text{EPS}=1\times128$}}
\\

Random 
& 30.1$_{\pm3.0}$ & 24.6$_{\pm3.1}$ 
& 35.8$_{\pm2.4}$ & 33.8$_{\pm4.1}$ 
& 31.1 
\\

Herding
& 33.7$_{\pm4.4}$ & 26.5$_{\pm2.1}$ 
& 41.3$_{\pm5.0}$ & 37.6$_{\pm2.3}$ 
& 34.8 
\\

DM
& 31.6$_{\pm3.3}$ & 26.5$_{\pm2.4}$ 
& 36.0$_{\pm1.4}$ & 35.6$_{\pm1.3}$ 
& 32.4 
\\
IDM 
& \textbf{41.1}$_{\pm2.6}$ & \textbf{32.1}$_{\pm2.5}$ 
& \underline{47.8}$_{\pm1.6}$ & \textbf{49.2}$_{\pm0.9}$ 
& \textbf{42.6} 
\\

SADM 
& 33.5$_{\pm3.2}$ & 28.7$_{\pm1.1}$ 
& 41.2$_{\pm5.9}$ & 41.1$_{\pm3.1}$ 
& 36.1 
\\

\rowcolor{MaDeSRow}MaDeS{} 
& \underline{37.1}$_{\pm4.6}$ & \underline{29.2}$_{\pm1.8}$ 
& \textbf{49.0}$_{\pm1.9}$ & \underline{45.2}$_{\pm2.1}$ 
& \underline{40.1} 
\\

\midrule

\multicolumn{6}{l}{\textbf{$\text{SPC}\times\text{EPS}=1\times256$}}
\\

Random 
& 26.5$_{\pm3.6}$ & 27.0$_{\pm2.6}$ 
& 35.3$_{\pm1.7}$ & 34.2$_{\pm2.3}$ 
& 30.8 
\\

Herding
& 31.0$_{\pm3.7}$ & \underline{33.5}$_{\pm3.9}$ 
& \underline{41.5}$_{\pm1.9}$ & 40.4$_{\pm2.4}$ 
& 36.4 
\\

DM
& 30.3$_{\pm2.1}$ & 26.9$_{\pm3.7}$ 
& 36.5$_{\pm6.0}$ & 38.2$_{\pm2.3}$ 
& 33.0 
\\

IDM 
& \textbf{39.1}$_{\pm3.5}$ & \textbf{34.1}$_{\pm2.2}$ 
& \textbf{44.7}$_{\pm2.8}$ & \underline{46.8}$_{\pm1.4}$ 
& \textbf{41.2} 
\\

SADM 
& 34.0$_{\pm2.3}$ & 30.7$_{\pm2.5}$ 
& 39.5$_{\pm5.6}$ & 43.2$_{\pm3.0}$ 
& 36.9 
\\

\rowcolor{MaDeSRow}MaDeS{} 
& \underline{36.5}$_{\pm3.2}$ & 32.7$_{\pm3.6}$ & 43.0$_{\pm2.9}$ 
& \textbf{47.3}$_{\pm2.4}$ 
& \underline{39.9} 
\\

\midrule

\multicolumn{6}{l}{\textbf{$\text{SPC}\times\text{EPS}=10\times256$}} 
\\

Random 
& 61.7$_{\pm1.8}$ & 49.4$_{\pm1.8}$ 
& 61.4$_{\pm4.5}$ & 69.0$_{\pm2.4}$ 
& 60.4 
\\

Herding
& 64.6$_{\pm2.3}$ & 51.0$_{\pm1.4}$ 
& 59.8$_{\pm2.6}$ & 70.9$_{\pm2.7}$ 
& 61.6 
\\

DM 
& 66.6$_{\pm1.8}$ & 52.7$_{\pm2.4}$ 
& 61.6$_{\pm2.3}$ & 74.3$_{\pm1.1}$ 
& 63.8 
\\

IDM 
& 71.7$_{\pm2.7}$ & \underline{59.2}$_{\pm2.2}$ 
& \underline{65.7}$_{\pm3.2}$ & \underline{77.3}$_{\pm1.4}$ 
& \underline{69.2} 
\\

SADM 
& \underline{71.8}$_{\pm2.8}$ & 53.4$_{\pm1.5}$ 
& 62.3$_{\pm2.4}$ & 77.2$_{\pm2.3}$ 
& 66.2 
\\

\rowcolor{MaDeSRow}MaDeS
& \textbf{78.5}$_{\pm1.7}$ & \textbf{60.4}$_{\pm2.5}$ 
& \textbf{66.6}$_{\pm1.0}$ & \textbf{82.1}$_{\pm0.9}$
& \textbf{71.9} 
\\

\bottomrule

\end{tabular}
\end{table}
\paragraph{Distillation and Evaluation.}
We summarize the optimization and evaluation settings in \Cref{tab:sup_dis_set}.
Specifically, during distillation, the real batches and synthetic samples receive the same random augmentation consisting of spatial scale in $[0.8,1.2]$, spatial and temporal shift of up to $\pm0.1$, temporal scale in $[0.8,1.2]$, Gaussian jitter $\sigma=0.01$ on $t,x,y$, and a global polarity flip with probability $0.1$.
During evaluation, the same optimization is used for train samples (\eg distilled samples) while no optimization is used for test samples.
On THU-EACT-50 and N-Caltech101, the classes are processed in chunks of 10 classes with 8 real samples per class per iteration to bound the GPU memory, which is equivalent to the unchunked loss in expectation.
All evaluation accuracies are computed on all test samples of each dataset, subsampled to the same EPS with a uniform stride.

\section{Algorithm}
We summarize the workflow of MaDeS in Algorithm~\ref{alg}. Firstly, classifiers are trained on real samples subsampled to the target EPS, and the checkpoints at the target maturity levels are collected to form the state pool. 
Then the learnable event parameters are initialized from real samples under the SPC and EPS budgets. 
At each iteration, a randomly initialized encoder extracts features for response-based distribution matching, while a classifier sampled from the state pool provides digits for cross-entropy calculation. 
After optimization, the event parameters are mapped back to raw events and sorted by timestamp to obtain the distilled event dataset.

\begin{table}[t]
\centering
\footnotesize
\setlength{\tabcolsep}{4pt}
\caption{Cross-architecture accuracy (\%) of sets distilled with SECNet at $\mathrm{SPC}\times\mathrm{EPS}=10\times128$.}
\label{tab:sup_cross_dataset}
\begin{tabular}{l|cccc|c}
\toprule
Method & SECNet & EventMamba & Voxel & SNN & Mean \\
\midrule

\multicolumn{6}{l}{\textit{N-MNIST}} 
\\

Random 
& 61.2$_{\pm2.1}$ & 49.8$_{\pm3.9}$
& 77.1$_{\pm1.4}$ & 74.1$_{\pm0.7}$
& 65.5 
\\

Herding
& 63.2$_{\pm1.3}$ & 54.3$_{\pm2.4}$ 
& \underline{81.9}$_{\pm1.5}$ & 76.0$_{\pm0.3}$ 
& 68.9 
\\

DM
& 70.1$_{\pm0.9}$ & 58.8$_{\pm4.3}$ 
& 80.3$_{\pm1.1}$ & 77.0$_{\pm1.1}$ 
& 71.6 
\\

IDM
& 69.4$_{\pm1.3}$ & \underline{70.8}$_{\pm1.2}$ 
& 77.6$_{\pm0.2}$ & 74.5$_{\pm0.4}$ 
& 73.1 
\\

SADM 
& \underline{72.2}$_{\pm0.8}$ & 62.0$_{\pm2.8}$ 
& 81.5$_{\pm0.7}$ & \underline{78.4}$_{\pm0.9}$ 
& \underline{73.5} 
\\

\rowcolor{MaDeSRow}MaDeS
& \textbf{76.9}$_{\pm1.0}$ & \textbf{72.3}$_{\pm3.5}$ 
& \textbf{83.8}$_{\pm0.3}$ & \textbf{81.2}$_{\pm0.3}$ 
& \textbf{78.5} 
\\

\midrule

\multicolumn{6}{l}{\textit{THU-EACT-50}} 
\\

Random 
& 13.9$_{\pm1.3}$ & 16.9$_{\pm1.6}$ 
& 15.5$_{\pm0.9}$ & 14.1$_{\pm0.8}$ 
& 15.1 
\\

Herding
& 15.2$_{\pm1.4}$ & 17.4$_{\pm1.1}$ 
& 17.3$_{\pm1.1}$ & 16.1$_{\pm0.9}$ 
& 16.5 
\\

DM 
& 15.6$_{\pm1.3}$ & 18.9$_{\pm0.8}$ 
& 16.5$_{\pm1.1}$ & 15.2$_{\pm0.9}$ 
& 16.5 
\\

IDM 
& \underline{23.7}$_{\pm1.8}$ & \underline{27.2}$_{\pm0.6}$ 
& \underline{23.7}$_{\pm0.7}$ & \underline{22.9}$_{\pm0.6}$ 
& \underline{24.4} 
\\

SADM 
& 17.3$_{\pm1.8}$ & 20.1$_{\pm1.3}$ 
& 18.3$_{\pm1.4}$ & 15.0$_{\pm1.1}$ 
& 17.7 
\\

\rowcolor{MaDeSRow}MaDeS
& \textbf{25.8}$_{\pm1.5}$ & \textbf{33.0}$_{\pm1.0}$ 
& \textbf{27.2}$_{\pm1.0}$ & \textbf{28.2}$_{\pm0.3}$ 
& \textbf{28.6} 
\\

\midrule

\multicolumn{6}{l}{\textit{N-Caltech101}} 
\\

Random 
& 13.6$_{\pm0.5}$ & 13.7$_{\pm1.8}$ 
& 25.6$_{\pm0.5}$ & 24.0$_{\pm0.6}$ 
& 19.2 
\\

Herding
& 14.6$_{\pm0.4}$ & 15.2$_{\pm1.3}$ 
& 26.7$_{\pm1.2}$ & 24.7$_{\pm0.7}$ 
& 20.3 
\\

DM 
& 16.9$_{\pm0.6}$ & 14.8$_{\pm1.0}$ 
& 25.7$_{\pm0.7}$ & 24.9$_{\pm0.5}$ 
& 20.6 
\\

IDM 
& \underline{22.8}$_{\pm0.7}$ & \underline{21.8}$_{\pm1.4}$ 
& \underline{32.1}$_{\pm0.4}$ & \underline{31.0}$_{\pm0.4}$ 
& \underline{27.0} 
\\

SADM 
& 17.8$_{\pm1.0}$ & 14.7$_{\pm1.1}$ 
& 25.7$_{\pm0.7}$ & 25.0$_{\pm0.3}$ 
& 20.8 
\\

\rowcolor{MaDeSRow}MaDeS & \textbf{26.9}$_{\pm0.9}$ & \textbf{28.2}$_{\pm1.1}$ 
& \textbf{36.8}$_{\pm0.4}$ & \textbf{35.2}$_{\pm0.4}$ 
& \textbf{31.8} 
\\

\bottomrule

\end{tabular}
\end{table}
\section{Cross-Architecture Generalization}
We extend the validation on the cross-architecture generalizability of MaDeS with more settings.
First, we evaluate SECNet-distilled datasets on unseen architectures on DVS128 Gesture under various distillation budgets.
As shown in \cref{tab:sup_cross_ipc}, individual evaluators occasionally favor different methods, but the mean follows the trend observed with SECNet. IDM leads at SPC$=1$, while MaDeS leads on every evaluator at the larger sample budget.
Then, we fix the budget at $\text{SPC}\times\text{EPS}=10\times128$ and repeat the comparison on other datasets. As shown in \cref{tab:sup_cross_dataset}, the mean rankings also vary across datasets. MaDeS keeps the first rank throughout at this distillation budget. SADM is the strongest baseline on N-MNIST, while IDM leads the baselines on the other two datasets.

\begin{table}[t]
\centering
\footnotesize
\setlength{\tabcolsep}{3.8pt}
\caption{Cross-architecture accuracy (\%) on DVS128 Gesture using EventMamba for distillation at $\mathrm{SPC}\times\mathrm{EPS}=10\times128$.}
\label{tab:supp_em}
\begin{tabular}{l|cccc|c}
\toprule
Method & SECNet & EventMamba & Voxel & SNN & Mean \\
\midrule
Random 
& 58.0$_{\pm1.4}$ & 43.5$_{\pm1.1}$ 
& 56.9$_{\pm4.1}$ & 66.7$_{\pm2.4}$ 
& 56.3 
\\

Herding 
& 61.1$_{\pm1.4}$ & 44.9$_{\pm1.6}$ 
& 56.7$_{\pm1.8}$ & 69.9$_{\pm0.6}$ 
& 58.1 
\\

DM~\cite{zhao2023dm} 
& 62.2$_{\pm1.6}$ & 50.5$_{\pm1.6}$ 
& 55.4$_{\pm1.8}$ & 71.2$_{\pm0.6}$ 
& 59.8 
\\

IDM~\cite{zhao2023idm} 
& \underline{67.3}$_{\pm1.2}$ & \textbf{57.7}$_{\pm2.0}$ 
& \underline{63.0}$_{\pm2.8}$ & 73.3$_{\pm1.4}$ 
& \underline{65.4} 
\\

SADM~\cite{yim2025sadm} 
& 65.6$_{\pm1.6}$ & 52.3$_{\pm2.6}$ 
& 58.3$_{\pm1.3}$ & \underline{74.8}$_{\pm1.1}$ 
& 62.7 
\\

\rowcolor{MaDeSRow}MaDeS{} 
& \textbf{70.0}$_{\pm1.0}$ & \underline{57.0}$_{\pm2.5}$ 
& \textbf{64.1}$_{\pm2.5}$ & \textbf{82.2}$_{\pm0.3}$ 
& \textbf{68.3} 
\\

\bottomrule

\end{tabular}
\end{table}
\begin{table}[t]
\centering
\footnotesize
\setlength{\tabcolsep}{4pt}
\caption{Ablation of the loss terms across datasets and distillation budgets. $\bigcirc$ denotes replacing $\mathcal{L}_{\mathrm{pres}}$ with standard matching of DM~\cite{zhao2023dm}. The best values per dataset and budget are highlighted in bold.}
\label{tab:sup_ablation}
\begin{tabular}{ll|cccc}
\toprule
$\mathcal{L}_\mathrm{pres}$ & $\mathcal{L}_\mathrm{disc}$
& 1$\times$128 & 1$\times$256 & 10$\times$128 & 10$\times$256
\\
\midrule
\multicolumn{6}{l}{\textbf{DVS128 Gesture}}
\\

$\bigcirc$ & $\times$
& $32.6_{\pm2.3}$ & $31.0_{\pm2.1}$ & $64.9_{\pm1.0}$ & $66.7_{\pm3.3}$
\\

$\times$ & $\checkmark$
& $32.0_{\pm3.2}$ & $28.8_{\pm2.8}$ & $66.2_{\pm1.3}$ & $63.3_{\pm3.2}$
\\

$\bigcirc$ & $\checkmark$
& $34.8_{\pm3.1}$ & $33.6_{\pm3.0}$ & $70.3_{\pm1.0}$ & $73.2_{\pm1.9}$
\\

$\checkmark$ & $\times$
& $33.7_{\pm3.4}$ & $34.4_{\pm2.6}$ & $68.3_{\pm0.9}$ & $71.0_{\pm2.5}$
\\

\rowcolor{MaDeSRow}
$\checkmark$ & $\checkmark$
& \bm{$37.1_{\pm4.6}$} & \bm{$36.5_{\pm3.2}$} & \bm{$73.4_{\pm1.1}$} & \bm{$78.5_{\pm1.7}$}
\\

\midrule

\multicolumn{6}{l}{\textbf{N-MNIST}}
\\

$\bigcirc$ & $\times$
& $38.1_{\pm3.6}$ & $43.1_{\pm2.5}$ & $69.4_{\pm0.9}$ & $82.3_{\pm0.8}$
\\

$\times$ & $\checkmark$
& $35.1_{\pm3.5}$ & $45.7_{\pm4.7}$ & $67.8_{\pm1.8}$ & $74.6_{\pm1.0}$
\\

$\bigcirc$ & $\checkmark$
& $41.6_{\pm2.2}$ & $50.6_{\pm3.7}$ & $72.8_{\pm1.8}$ & $82.7_{\pm0.7}$
\\

$\checkmark$ & $\times$
& $44.0_{\pm3.6}$ & $54.7_{\pm5.3}$ & $72.1_{\pm0.9}$ & $84.9_{\pm1.0}$
\\

\rowcolor{MaDeSRow}
$\checkmark$ & $\checkmark$
& \bm{$45.5_{\pm2.5}$} & \bm{$57.2_{\pm4.1}$} & \bm{$76.9_{\pm1.0}$} & \bm{$87.2_{\pm0.5}$}
\\

\midrule

\multicolumn{6}{l}{\textbf{THU-EACT-50}}
\\

$\bigcirc$ & $\times$
& $10.5_{\pm1.3}$ & $9.0_{\pm0.7}$ & $16.2_{\pm1.6}$ & $21.6_{\pm1.2}$
\\

$\times$ & $\checkmark$
& $8.2_{\pm1.4}$ & $9.8_{\pm1.6}$ & $20.1_{\pm1.2}$ & $26.7_{\pm1.9}$
\\

$\bigcirc$ & $\checkmark$
& $11.9_{\pm1.7}$ & $12.4_{\pm1.0}$ & $23.8_{\pm1.1}$ & $34.7_{\pm1.2}$
\\

$\checkmark$ & $\times$
& $12.5_{\pm1.2}$ & $12.2_{\pm1.3}$ & $17.4_{\pm1.4}$ & $25.1_{\pm1.3}$
\\

\rowcolor{MaDeSRow}
$\checkmark$ & $\checkmark$
& \bm{$13.6_{\pm2.3}$} & \bm{$13.7_{\pm1.9}$} & \bm{$25.8_{\pm1.5}$} & \bm{$39.6_{\pm2.2}$}
\\

\midrule
\multicolumn{6}{l}{\textbf{N-Caltech101}}
\\

$\bigcirc$ & $\times$
& $8.5_{\pm2.0}$ & $9.4_{\pm1.6}$ & $17.2_{\pm0.7}$ & $23.1_{\pm0.7}$
\\

$\times$ & $\checkmark$
& $9.1_{\pm1.3}$ & $11.0_{\pm1.4}$ & $24.0_{\pm0.7}$ & $30.7_{\pm0.8}$
\\

$\bigcirc$ & $\checkmark$
& $12.4_{\pm0.7}$ & $13.4_{\pm2.1}$ & $26.4_{\pm0.7}$ & $33.0_{\pm0.8}$
\\

$\checkmark$ & $\times$
& $9.5_{\pm2.2}$ & $12.1_{\pm1.0}$ & $17.5_{\pm0.7}$ & $24.7_{\pm0.8}$
\\

\rowcolor{MaDeSRow}
$\checkmark$ & $\checkmark$
& \bm{$13.3_{\pm0.9}$} & \bm{$16.2_{\pm0.9}$} & \bm{$26.9_{\pm0.9}$} & \bm{$34.4_{\pm0.9}$}
\\

\bottomrule
\end{tabular}
\end{table}
\section{Transferability on Distillation Model}
We replace SECNet with EventMamba~\cite{ren2024eventmamba} as the distillation network and compare the accuracy with baselines on DVS128 Gesture under $\mathrm{SPC}\times\mathrm{EPS}=10\times128$.
The baselines also use EventMamba as the distillation model or feature extractor.
As shown in \cref{tab:supp_em}, IDM has a small lead when evaluated with EventMamba itself. This advantage does not transfer to the other architectures, where MaDeS consistently achieves higher accuracy. And MaDeS achieves the best mean accuracy under the evaluated setting.

\section{Sensitivity to \texorpdfstring{$\lambda$}{lambda} in Other Settings}

\begin{figure*}[t]
\centering
\includegraphics[width=16cm]{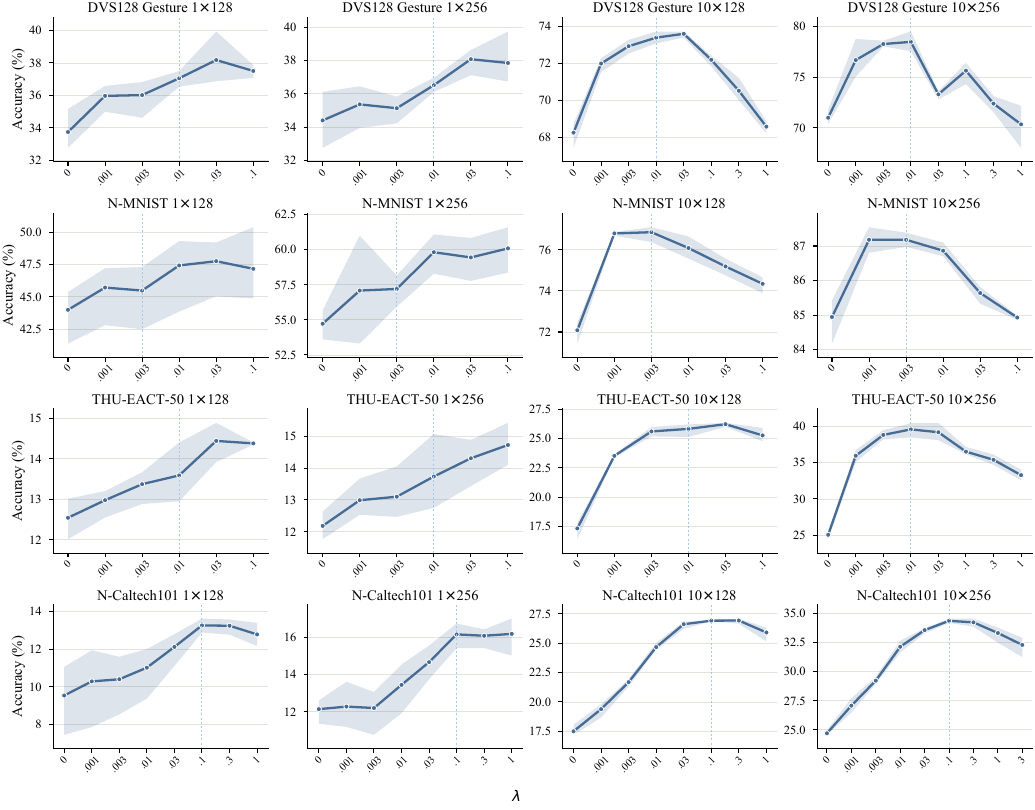}
\caption{Accuracy (\%) across discrimination weights $\lambda$ for additional datasets and distillation budgets.}
\label{fig:supp_lambda}
\end{figure*}

We further validate the sensitivity of MaDeS to $\lambda$ by varying the weight term $\lambda$ across different datasets and distillation budgets.
As shown in \cref{fig:supp_lambda}, curves average over distilled sets and training seeds, bands span the per-set means, and dashed markers indicate the dataset-specific defaults. 
A modest discrimination weight generally improves accuracy over matching alone. On Gesture, the curves peak at smaller weights when more samples per class are available. With only one sample per class, accuracy continues to improve over a wider range before leveling off, although variation across sets is also larger. This shift is consistent with discrimination becoming more important when each class has few representatives. The defaults selected at the larger sample budget lie near the peaks there and still improve on matching alone at the smaller Gesture budgets.

\section{Ablation on Distillation Objective on Other Settings}
\begin{table}[t]
\renewcommand{\MaDeSMaxGain}{29.0}
\centering
\footnotesize
\setlength{\tabcolsep}{4pt}
\caption{Accuracy (\%) of MaDeS on N-MNIST and THU-EACT-50 across distillation budgets. Parentheses give differences from Herding in percentage points. Darker shading indicates larger gains.}
\label{tab:supp_grid}
\begin{tabular}{c|cccc}
\toprule
\multicolumn{5}{c}{N-MNIST} \\
SPC$\backslash$EPS & 128 & 256 & 512 & 1024 \\
\midrule
1
& \gaincell{45.5}{+17.3}
& \gaincell{57.2}{+16.9}
& \gaincell{51.2}{+12.1}
& \gaincell{36.1}{+8.0}
\\

5
& \gaincell{70.5}{+16.4}
& \gaincell{83.7}{+11.6}
& \gaincell{88.2}{+5.1}
& \gaincell{86.7}{+6.0}
\\

10
& \gaincell{76.9}{+13.7}
& \gaincell{87.2}{+5.4}
& \gaincell{91.4}{+2.7}
& \gaincell{93.8}{+2.7}
\\

20
& \gaincell{79.3}{+7.2}
& \gaincell{88.9}{+2.5}
& \gaincell{92.8}{+1.0}
& \gaincell{94.9}{+0.3}
\\

50
& \gaincell{81.2}{+1.7}
& \gaincell{89.8}{-0.8}
& \gaincell{93.6}{-1.1}
& \gaincell{95.4}{-1.3}
\\

\midrule

\multicolumn{5}{c}{THU-EACT-50} \\
SPC$\backslash$EPS & 128 & 256 & 512 & 1024 \\
\midrule
1
& \gaincell{13.6}{+6.8}
& \gaincell{13.7}{+6.3}
& \gaincell{11.3}{+5.1}
& \gaincell{9.4}{+3.7}
\\

5
& \gaincell{22.0}{+9.6}
& \gaincell{31.0}{+16.4}
& \gaincell{42.3}{+25.7}
& \gaincell{48.8}{+29.0}
\\

10
& \gaincell{25.8}{+10.6}
& \gaincell{39.6}{+19.3}
& \gaincell{49.9}{+23.3}
& \gaincell{56.0}{+23.2}
\\

20
& \gaincell{28.2}{+9.9}
& \gaincell{42.8}{+17.0}
& \gaincell{51.7}{+12.6}
& \gaincell{57.5}{+8.3}
\\

50
& \gaincell{29.2}{+3.2}
& \gaincell{43.2}{+6.3}
& \gaincell{52.9}{+2.0}
& \gaincell{59.2}{-1.5}
\\
\bottomrule
\end{tabular}
\end{table}
We further isolate the contribution of preservation and discrimination losses across datasets and distillation budgets.
As shown in \cref{tab:sup_ablation}, the results are consistent with the main experiments. Adding response-based matching improves classification accuracy, and adding the cross-entropy loss yields further gains. The relative contributions depend on the number of samples per class and dataset. At low SPC, response-based matching accounts for the larger gain, whereas at high SPC, the additional gain from CE is more pronounced. On N-Caltech101 with the most number of classes, the discrimination term starts to count even at low EPS settings.

\section{Two-Axis Compression on Other Datasets}
We extend evaluation under more distillation budgets to more datasets and compare MaDeS with the selection baseline Herding. As shown in \cref{tab:supp_grid}, the budget at which MaDeS achieves its largest gain over Herding varies by dataset. 
On N-MNIST, the largest gain occurs when both SPC and EPS are low, suggesting that relatively few samples and events may suffice to capture the features needed for digit recognition. 
On THU-EACT-50, MaDeS gains the most at moderate SPC and higher EPS, a pattern also observed on DVS128 Gesture in the main paper. 
As SPC increases, Herding generally narrows the gap and can slightly outperform MaDeS when both SPC and EPS are high.

\section{Combination of Selection Methods}
\label{sec:selection}
The selection baselines in the main paper apply a selection criterion along the target axis and use random sampling along the other axis. 
We further combine sample and event selection and compare the performance with MaDeS on DVS128 Gesture across distillation budgets. 
We evaluate both orders because subsampling events first can change which samples a criterion such as Herding selects. 
In \cref{tab:sup_selection}, we first select samples within each class and then subsample events within each selected sample. 
And we subsample events before selecting samples in \cref{tab:sup_selection_reverse}.

The results show that applying a selection criterion to each axis can improve accuracy, but the gains are inconsistent. 
At some budgets, using random sampling on one or both axes yields higher accuracy than selecting along both axes. 
The best-performing selection combination also varies with the distillation budgets and can change when the selection order is reversed. 
MaDeS outperforms every tested selection combination at all evaluated settings under both selection orders.

\begin{table}[t]
\centering
\scriptsize
\setlength{\tabcolsep}{2pt}
\caption{Accuracy (\%) of sample and event selection combinations on DVS128 Gesture across SPC $\times$ EPS budgets. Sample selection is conducted before event selection.}
\label{tab:sup_selection}
\begin{tabular}{ll|cccc}
\toprule

Sample & Event 
& 1$\times$128 & 1$\times$256 
& 10$\times$128 & 10$\times$256 
\\

\midrule

\multirow{3}{*}{Random} & Random
& 30.1$_{\pm3.0}$ & 26.5$_{\pm3.6}$ 
& 58.0$_{\pm1.4}$ & 61.7$_{\pm1.8}$ 
\\

& Stride
& \underline{33.8}$_{\pm2.8}$ & 31.9$_{\pm3.2}$ 
& 58.0$_{\pm2.3}$ & 61.4$_{\pm3.2}$ 
\\

& Araghi \etal
& 26.7$_{\pm3.0}$ & 26.5$_{\pm3.2}$ 
& 57.4$_{\pm1.8}$ & 59.1$_{\pm3.5}$ 
\\

& POLED
& 27.3$_{\pm3.6}$ & 27.5$_{\pm2.3}$ 
& 55.3$_{\pm2.9}$ & 59.6$_{\pm2.9}$ 
\\

\midrule
\multirow{3}{*}{Herding} & Random
& 33.7$_{\pm4.4}$ & 31.0$_{\pm3.7}$ 
& 61.7$_{\pm0.8}$ & 64.6$_{\pm2.3}$ 
\\ 

& Stride
& 30.9$_{\pm2.5}$ & 31.4$_{\pm3.2}$ 
& \underline{62.9}$_{\pm1.6}$ & 65.6$_{\pm2.2}$ 
\\ 

& Araghi \etal
& 31.4$_{\pm3.5}$ & \underline{33.6}$_{\pm2.7}$ 
& 61.4$_{\pm1.3}$ & \underline{65.9}$_{\pm2.3}$ 
\\

& POLED
& 30.3$_{\pm4.4}$ & 31.4$_{\pm3.2}$ 
& 59.6$_{\pm1.2}$ & 64.1$_{\pm2.8}$ 
\\

\midrule

\multirow{3}{*}{Closest} & Random
& 33.4$_{\pm4.4}$ & 31.0$_{\pm3.1}$
& 57.1$_{\pm2.2}$ & 58.8$_{\pm3.1}$ 
\\

& Stride
& 32.1$_{\pm2.6}$ & 30.8$_{\pm3.2}$ 
& 58.5$_{\pm0.8}$ & 58.8$_{\pm2.5}$
\\ 

& Araghi \etal
& 31.7$_{\pm3.6}$ & 33.4$_{\pm1.9}$ 
& 56.4$_{\pm2.1}$ & 56.6$_{\pm2.6}$ 
\\

& POLED
& 30.3$_{\pm3.9}$ & 31.6$_{\pm2.8}$ 
& 54.0$_{\pm2.0}$ & 56.6$_{\pm1.9}$ 
\\

\midrule
\rowcolor{MaDeSRow}\multicolumn{2}{c|}{MaDeS} 
& \bm{$37.1_{\pm4.6}$} & \bm{$36.5_{\pm3.2}$} 
& \bm{$73.4_{\pm1.1}$} & \bm{$78.5_{\pm1.7}$}
\\
\bottomrule
\end{tabular}
\end{table}
\begin{table}[t]
\centering
\scriptsize
\setlength{\tabcolsep}{2pt}
\caption{Accuracy (\%) of sample and event selection combinations on DVS128 Gesture across distillation budgets. Event selection is conducted before sample selection.}
\label{tab:sup_selection_reverse}
\begin{tabular}{ll|cccc}
\toprule
Event & Sample & 1$\times$128 & 1$\times$256 & 10$\times$128 & 10$\times$256 \\
\midrule
\multirow{3}{*}{Random} & Random
& 30.1$_{\pm3.0}$ & 26.5$_{\pm3.6}$ & 58.0$_{\pm1.4}$ & 61.7$_{\pm1.8}$ \\
 & Herding
& \underline{34.5}$_{\pm2.3}$ & \underline{33.6}$_{\pm4.7}$ & 62.3$_{\pm2.1}$ & 63.9$_{\pm2.5}$
\\
 & Closest
& 34.4$_{\pm2.2}$ & \underline{33.6}$_{\pm3.9}$ & 57.3$_{\pm1.2}$ & 59.4$_{\pm2.6}$
\\
\midrule
\multirow{3}{*}{Stride} & Random
& 33.8$_{\pm2.8}$ & 31.9$_{\pm3.2}$ & 58.0$_{\pm2.3}$ & 61.4$_{\pm3.2}$ \\
 & Herding
& 31.6$_{\pm2.3}$ & 31.9$_{\pm2.7}$ & \underline{62.6}$_{\pm0.9}$ & \underline{66.7}$_{\pm1.7}$
\\
 & Closest
& 31.4$_{\pm2.7}$ & 32.7$_{\pm2.0}$ & 57.1$_{\pm0.8}$ & 58.7$_{\pm1.6}$
\\
\midrule
\multirow{3}{*}{Araghi \etal} & Random
& 26.7$_{\pm3.0}$ & 26.5$_{\pm3.2}$ & 57.4$_{\pm1.8}$ & 59.1$_{\pm3.5}$ \\
 & Herding
& 32.0$_{\pm3.3}$ & 30.8$_{\pm3.4}$ & 60.6$_{\pm1.8}$ & 65.1$_{\pm3.5}$
\\
 & Closest
& 31.7$_{\pm3.0}$ & 30.9$_{\pm2.4}$ & 57.6$_{\pm1.5}$ & 56.1$_{\pm3.0}$
\\
\midrule
\multirow{3}{*}{POLED} & Random
& 27.3$_{\pm3.6}$ & 27.5$_{\pm2.3}$ & 55.3$_{\pm2.9}$ & 59.6$_{\pm2.9}$ \\
 & Herding
& 29.6$_{\pm2.0}$ & 30.3$_{\pm3.0}$ & 60.5$_{\pm2.9}$ & 64.8$_{\pm2.1}$
\\
 & Closest
& 29.6$_{\pm2.3}$ & 30.6$_{\pm2.8}$ & 57.0$_{\pm2.3}$ & 55.8$_{\pm1.5}$
\\
\midrule
\rowcolor{MaDeSRow}\multicolumn{2}{c|}{MaDeS}
& \bm{$37.1_{\pm4.6}$} & \bm{$36.5_{\pm3.2}$} & \bm{$73.4_{\pm1.1}$} & \bm{$78.5_{\pm1.7}$}
\\
\bottomrule
\end{tabular}
\end{table}

\section{Sensitivity to Composition of the State Pool}
\label{sec:sup_state_pool}
We examine the effect of state pool maturity on MaDeS by varying selected checkpoints from different training stages.
Early and Late use the first and last four saved checkpoints per trajectory, respectively. 
Uniform selects four checkpoints evenly spaced in training iterations, and MaDeS selects checkpoints by maturity.

\par

As shown in \Cref{tab:supp_pool}, DVS128 Gesture generally favors early states, whereas N-MNIST tends to favor late states. THU-EACT-50 shows no consistent preference for either stage. 
The uniform pool performs similarly to MaDeS overall (within the standard deviation), suggesting that once the pool covers a range of training stages, performance is relatively insensitive to the exact checkpoints included.
MaDeS performs more consistently under the evaluated settings.

\begin{table}[t]
\centering
\footnotesize
\caption{Accuracy (\%) for different classifier state pools at $\mathrm{SPC}\times\mathrm{EPS}=10\times128$.}
\label{tab:supp_pool}
\begin{tabular}{l|ccc}
\toprule
Pool & Gesture & N-MNIST & THU-EACT-50 
\\
\midrule
\multicolumn{4}{l}{\textbf{SPC $\times$ EPS $=$ 1$\times$128}} \\

Early & \bm{$37.6_{\pm3.5}$} & \bm{$46.8_{\pm4.4}$} & $13.9_{\pm2.1}$ \\
Late & $35.5_{\pm3.5}$ & $45.5_{\pm3.4}$ & $13.2_{\pm2.3}$ \\
Uniform & $36.2_{\pm3.8}$ & $45.0_{\pm4.1}$ & \bm{$14.0_{\pm2.2}$} \\
\rowcolor{MaDeSRow}MaDeS & $37.1_{\pm4.6}$ & $45.5_{\pm2.5}$ & $13.6_{\pm2.3}$ \\
\midrule

\multicolumn{4}{l}{\textbf{SPC $\times$ EPS $=$ 1$\times$256}} 
\\

Early & \bm{$38.1_{\pm3.9}$} & $57.0_{\pm4.5}$ & $13.2_{\pm1.6}$ \\
Late & $35.4_{\pm1.9}$ & $55.6_{\pm5.7}$ & $12.5_{\pm1.6}$ \\
Uniform & $36.6_{\pm3.2}$ & $56.3_{\pm2.7}$ & $13.2_{\pm1.7}$ \\
\rowcolor{MaDeSRow}MaDeS & $36.5_{\pm3.2}$ & \bm{$57.2_{\pm4.1}$} & \bm{$13.7_{\pm1.9}$} 
\\
\midrule

\multicolumn{4}{l}{\textbf{SPC $\times$ EPS $=$ 10$\times$128}} \\
Early & $73.2_{\pm0.9}$ & $75.7_{\pm1.3}$ & $25.2_{\pm1.4}$ \\
Late & $70.7_{\pm1.1}$ & $76.9_{\pm0.5}$ & $25.3_{\pm2.8}$ \\
Uniform & $72.7_{\pm1.0}$ & $76.6_{\pm0.8}$ & \bm{$26.3_{\pm1.7}$} \\
\rowcolor{MaDeSRow}MaDeS & \bm{$73.4_{\pm1.1}$} & \bm{$76.9_{\pm1.0}$} & $25.8_{\pm1.5}$ \\

\midrule

\multicolumn{4}{l}{\textbf{SPC $\times$ EPS $=$ 10$\times$256}} \\
Early & $77.6_{\pm0.9}$ & $86.6_{\pm0.9}$ & $37.3_{\pm1.3}$ \\
Late & $74.7_{\pm2.9}$ & \bm{$87.8_{\pm0.3}$} & $37.9_{\pm2.9}$ \\
Uniform & $77.9_{\pm1.5}$ & $87.5_{\pm0.5}$ & $39.4_{\pm2.5}$ \\
\rowcolor{MaDeSRow}MaDeS & \bm{$78.5_{\pm1.7}$} & $87.2_{\pm0.5}$ & \bm{$39.6_{\pm2.2}$} \\

\bottomrule
\end{tabular}
\end{table}

\begin{table}[t]
\centering
\footnotesize
\setlength{\tabcolsep}{3pt}
\caption{Accuracy (\%) for different matching objectives, matching networks, and classification supervision in IDM and SADM ablations.}
\label{tab:supp_idm_decomp}
\begin{tabular}{lc|cccc}
\toprule
Dataset & SPC$\times$EPS & DM & DM with Queue & IDM
\\

\midrule

Gesture & 1$\times$128
& 31.6$_{\pm3.3}$ & 35.3$_{\pm3.0}$ 
& 41.1$_{\pm2.6}$
\\

Gesture & 10$\times$128 & 64.9$_{\pm1.0}$ & 65.8$_{\pm1.5}$ & 69.7$_{\pm1.1}$
\\

N-MNIST & 1$\times$128 & 40.3$_{\pm2.3}$ & 40.8$_{\pm3.1}$ & 46.0$_{\pm3.9}$ 
\\

N-MNIST & 10$\times$128 & 70.1$_{\pm0.9}$ & 69.2$_{\pm1.6}$ & 69.4$_{\pm1.3}$ 
\\


THU-EACT-50 & 1$\times$128 
& 10.6$_{\pm1.2}$ & 8.5$_{\pm0.6}$ 
& 8.7$_{\pm0.8}$
\\

THU-EACT-50 & 10$\times$128 
& 15.6$_{\pm1.3}$ & 15.1$_{\pm1.8}$ 
& 23.7$_{\pm1.8}$ 
\\

\bottomrule
\end{tabular}
\end{table}
\begin{table}[t]
\centering
\footnotesize
\setlength{\tabcolsep}{2pt}
\caption{Training steps and classifier maturity (accuracy in \%) for the IDM queue and MaDeS state pool. Steps are reported as the median with the 10th and 90th percentiles in parentheses.}
\label{tab:supp_idm_maturity}
\begin{tabular}{lc|cc|cc}
\toprule
\multirow{2}{*}{Dataset} & \multirow{2}{*}{SPC$\times$EPS} 
& \multicolumn{2}{c|}{IDM} & \multicolumn{2}{c}{MaDeS} 
\\

& & Maturity & Accuracy 
& Maturity & Accuracy 
\\

\midrule

Gesture & 1$\times$128 
& 39.2 & $41.1_{\pm2.6}$ 
& 90.8 & $37.1_{\pm4.6}$
\\

Gesture & 10$\times$128  
& 84.6 & $69.7_{\pm1.1}$
& 90.8 & $73.4_{\pm1.1}$
\\

N-MNIST & 1$\times$128 
& 54.3 & $46.0_{\pm3.9}$
& 86.0 & $45.5_{\pm2.5}$
\\

N-MNIST & 10$\times$128 
& 80.8 & $69.4_{\pm1.2}$
& 86.0 & $76.9_{\pm1.0}$
\\


THU-EACT-50 & 1$\times$128  
& 7.6 & $8.7_{\pm0.8}$ 
& 46.0 & $13.6_{\pm2.3}$
\\

THU-EACT-50 & 10$\times$128  
& 30.7 & $23.7_{\pm1.8}$ 
& 46.0 & $25.8_{\pm1.5}$
\\
\bottomrule
\end{tabular}
\end{table}
\begin{table}[t]
\centering
\footnotesize
\setlength{\tabcolsep}{3pt}
\caption{IDM with more training steps per queue member and iteration. The default value is 10.}
\label{tab:supp_idm_steps}
\begin{tabular}{lc|ccc|cc}
\toprule
Dataset & Setting & 10 & 30 & 50 & MaDeS \\
\midrule
\multirow{2}{*}{Gesture} & 1$\times$128 & 41.1$_{\pm2.6}$ & 40.7$_{\pm4.9}$ & 40.5$_{\pm3.7}$ & 37.1$_{\pm4.6}$ \\
& 10$\times$128 & 69.7$_{\pm1.1}$ & 70.2$_{\pm1.2}$ & 69.8$_{\pm1.0}$ & 73.4$_{\pm1.1}$ \\

\midrule

\multirow{2}{*}{THU-EACT-50} & 1$\times$128 & 8.7$_{\pm0.8}$ & 9.8$_{\pm1.6}$ & 10.1$_{\pm1.2}$ & 13.6$_{\pm2.3}$ 
\\

& 10$\times$128 & 23.7$_{\pm1.8}$ & 24.4$_{\pm1.9}$ & 24.7$_{\pm1.7}$ & 25.8$_{\pm1.5}$ \\
\bottomrule

\end{tabular}
\end{table}

\begin{figure*}[t]
\centering
\includegraphics[width=16cm]{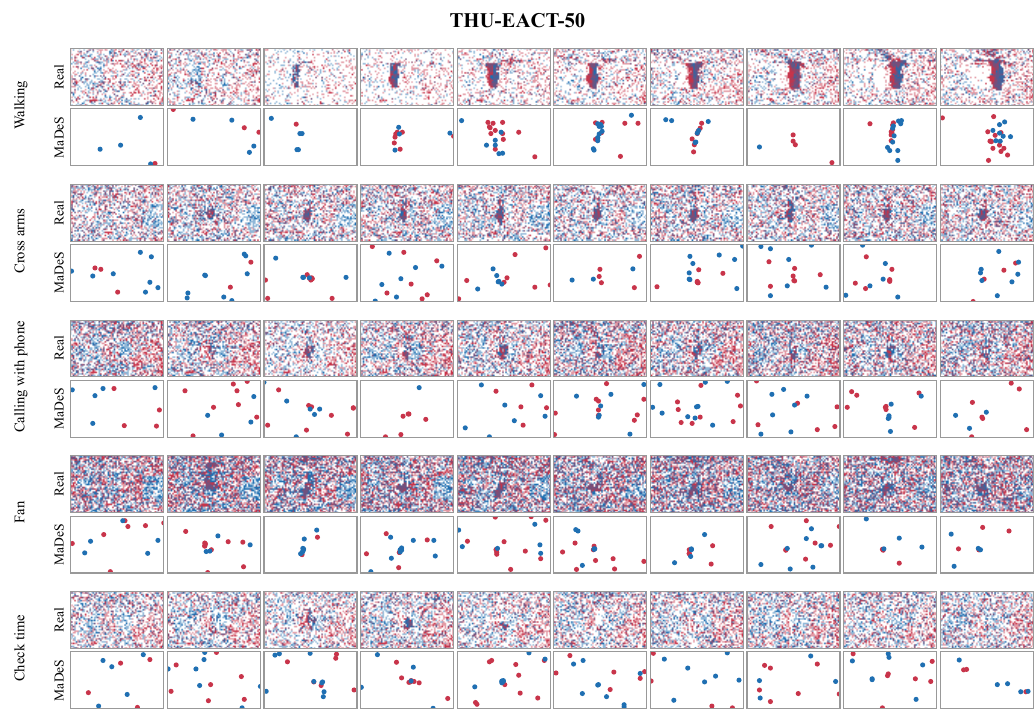}
\caption{Real and distilled event samples across time bins on THU-EACT-50. Red and blue indicate ON and OFF events, respectively.}
\label{fig:sup_vi_thu}
\end{figure*}

\begin{figure*}[p]
\centering
\includegraphics[width=\textwidth]{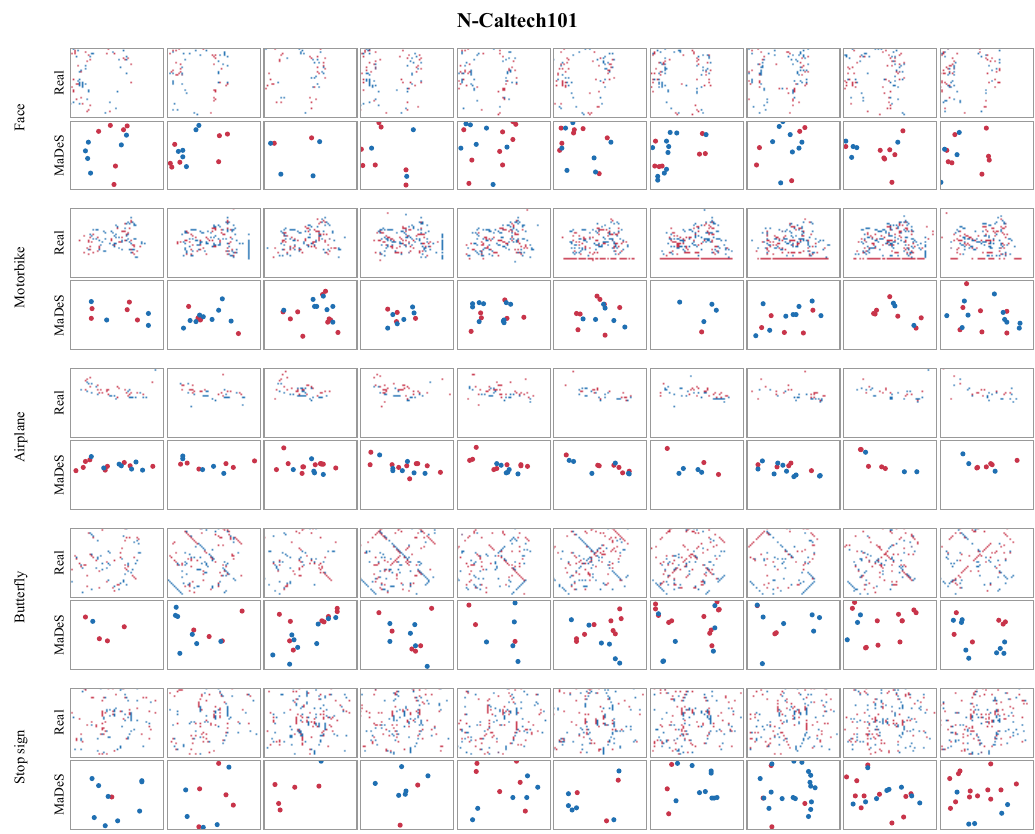}
\caption{Real and distilled event samples across time bins on N-Caltech101. Red and blue indicate ON and OFF events, respectively.}
\label{fig:sup_vi_nc101}
\end{figure*}

\section{Further Analysis on IDM}
\label{sec:supp_idm_sadm}
IDM is a strong baseline in our main experiments, but its performance varies considerably across settings. 
In the main paper, we attributed this variation in part to the maturity of the models in its online pool. 
Here, we provide detailed discussions on the contribution of model maturity for matching and discrimination and the inherent preference under different settings.
At each update, IDM selects one model from the pool and trains it for 10 steps. It regularly adds newly initialized models and removes the oldest model whenever the pool size exceeds 100. Each model therefore has a limited window in which it can accumulate training steps, and its maturity at removal depends on how often it was selected during that window.

\par

Firstly, we compare standard DM, using IDM online models for DM matching, and IDM. We also report the maturity of IDM models compared with MaDeS. As shown in \cref{tab:supp_idm_decomp}, switching to the online pool for matching yields more consistent changes, while the gain from cross-entropy loss varies more substantially.
As shown in \cref{tab:supp_idm_maturity}, models in IDM’s online pool generally have low maturity. With the preference discussed in \cref{sec:sup_state_pool}, IDM performs better in settings that favor less mature models. In settings that favor more mature models, the gain from CE is smaller and can even become negative.

\par

To test this interpretation, we increase the number of training steps per model update in IDM. As shown in \cref{tab:supp_idm_steps}, increasing model maturity reduces distillation accuracy on Gesture but slightly improves it on THU. These results support our interpretation that the effectiveness of IDM depends in part on model maturity and that the preferred level of maturity varies across settings.

\section{Visualization.}
\begin{figure*}[p]
\centering
\includegraphics[width=\textwidth]{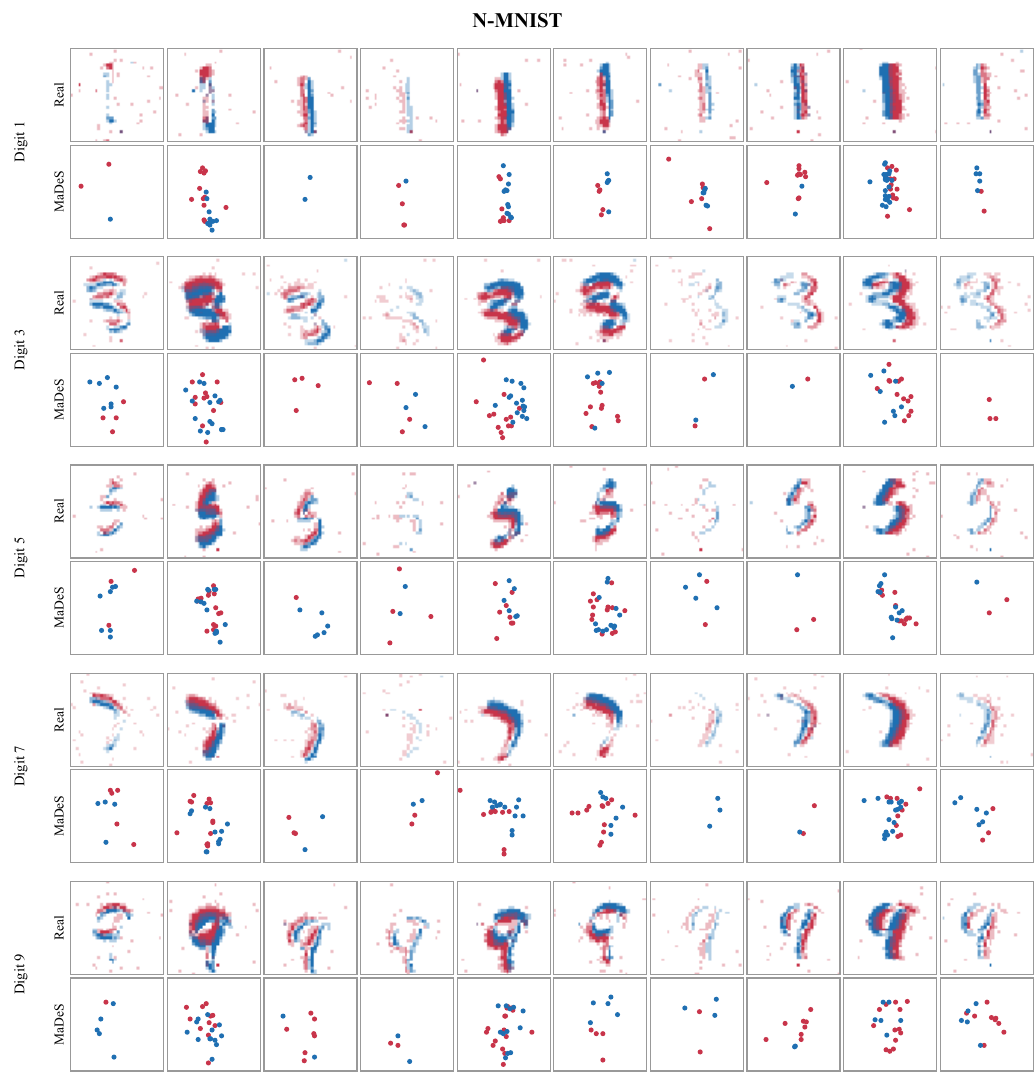}
\caption{Real and distilled event samples across time bins on N-MNIST. Red and blue indicate ON and OFF events, respectively.}
\label{fig:sup_vi_nmnist}
\end{figure*}

\begin{figure*}[p]
\centering
\includegraphics[width=\textwidth]{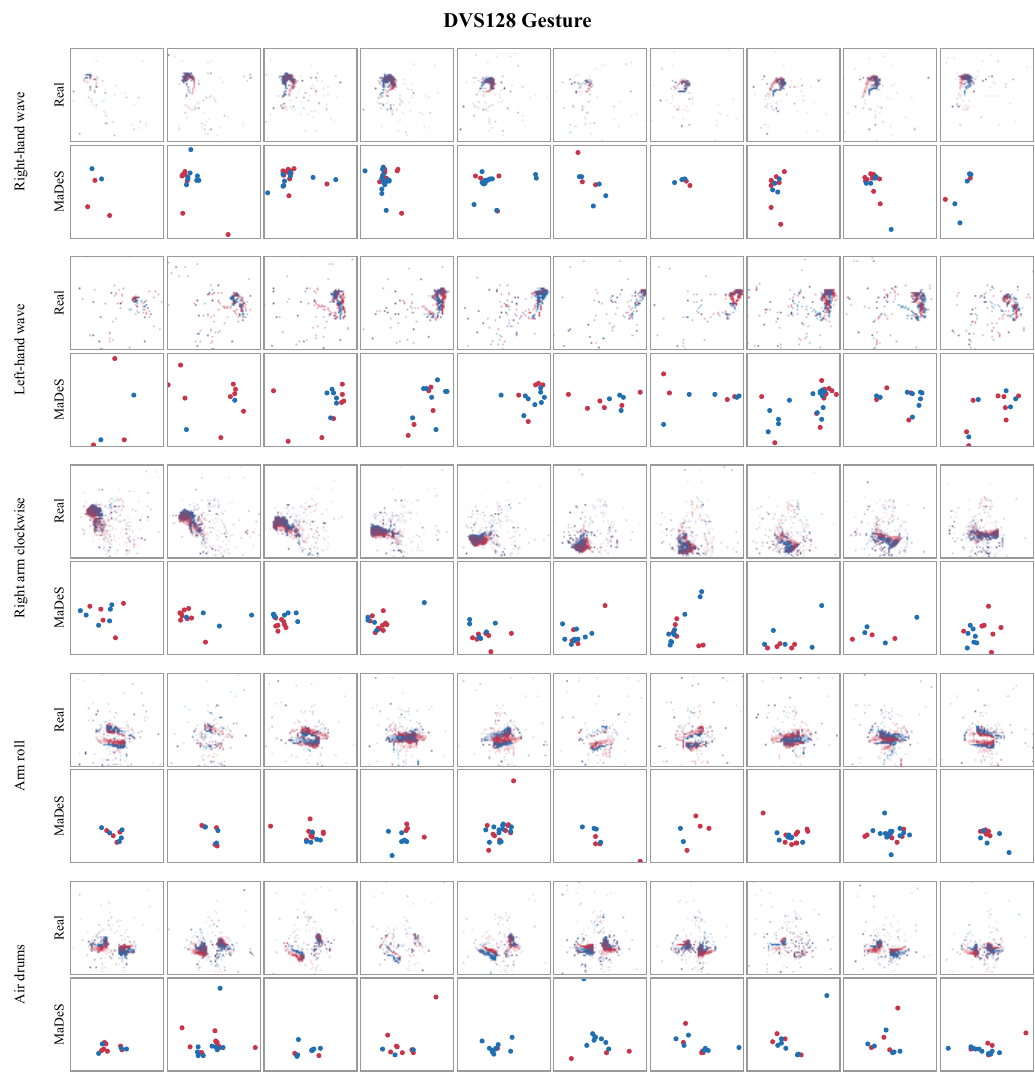}
\caption{Real and distilled event samples across time bins on DVS128 Gesture. Red and blue indicate ON and OFF events, respectively.}
\label{fig:sup_vi_gesture}
\end{figure*}

For each dataset, we select five classes and show one distilled sample per class alongside the real sample used for initialization. We divide each pair into the same consecutive time bins so that changes in event activity can be compared over time. \Cref{fig:sup_vi_thu,fig:sup_vi_nc101,fig:sup_vi_nmnist,fig:sup_vi_gesture} show that the distilled samples are much sparser than their source recordings. Even at this reduced EPS, the occupied regions broadly follow the temporal progression of activity in the real samples.

\end{document}